%% file: acmtog-main.tex
\documentclass[acmtog]{acmart}

\makeatletter
\def\@ACM@checkaffil{
    \if@ACM@instpresent\else
    \ClassWarningNoLine{\@classname}{No institution present for an affiliation}%
    \fi
    \if@ACM@citypresent\else
    \ClassWarningNoLine{\@classname}{No city present for an affiliation}%
    \fi
    \if@ACM@countrypresent\else
        \ClassWarningNoLine{\@classname}{No country present for an affiliation}%
    \fi
}
\makeatother

\AtBeginDocument{%
  }

\setcopyright{acmcopyright}
\copyrightyear{2023}
\acmYear{2023}
\acmDOI{XXXXXXX.XXXXXXX}

\usepackage{bbm}
\usepackage{amsmath}
\usepackage[ruled]{algorithm2e} 

\SetAlFnt{\small}
\SetAlCapFnt{\small}
\SetAlCapNameFnt{\small}
\SetAlCapHSkip{0pt}

\acmJournal{TOG}

\usepackage{multirow}
\usepackage[capitalize]{cleveref}
\crefname{section}{Sec.}{Secs.}
\crefname{table}{Tab.}{Tabs.}

\input{symbols}

\begin{document}

\title{\emph{EMS}: 3D Eyebrow Modeling from Single-view Images}

\author{Chenghong Li}
\authornote{Both authors contributed equally to this work.}
\email{chenghongli@link.cuhk.edu.cn}
\orcid{0009-0004-0604-7421}
\affiliation{%
  \institution{FNii and SSE, CUHKSZ}
  \city{Shenzhen}
  \country{China}
}

\author{Leyang Jin}
\authornotemark[1]
\email{leyangjin1@link.cuhk.edu.cn}
\orcid{0009-0002-1440-9096}
\affiliation{%
  \institution{SSE, CUHKSZ}
  \city{Shenzhen}
  \country{China}
}

\author{Yujian Zheng}
\orcid{0000-0001-7784-8323}
\email{yujianzheng@link.cuhk.edu.cn}
\affiliation{%
  \institution{FNii and SSE, CUHKSZ}
  \city{Shenzhen}
  \country{China}
}

\author{Yizhou Yu}
\authornotemark[2]
\orcid{0000-0002-0470-5548}
\email{yizhouy@acm.org}
\affiliation{%
  \institution{The University of Hong Kong}
}

\author{Xiaoguang Han}
\authornote{Corresponding author}
\orcid{0000-0003-0162-3296}
\email{hanxiaoguang@cuhk.edu.cn}
\affiliation{%
  \institution{SSE and FNii, CUHKSZ}
  \city{Shenzhen}
  \country{China}
}

\input{figure/fig_teaser}

\begin{abstract}
Eyebrows play a critical role in facial expression and appearance. Although the 3D digitization of faces is well explored, less attention has been drawn to 3D eyebrow modeling. In this work, we propose \ems, the first learning-based framework for single-view 3D eyebrow reconstruction. Following the methods of scalp hair reconstruction, we also represent the eyebrow as a set of fiber curves and convert the reconstruction to fibers growing problem. Three modules are then carefully designed: \rootFinder firstly localizes the fiber root positions which indicate where to grow; \oriPredictor predicts an orientation field in the 3D space to guide the growing of fibers; \fiberEnder is designed to determine when to stop the growth of each fiber. Our \oriPredictor directly borrows the method used in hair reconstruction. Considering the differences between hair and eyebrows, both \rootFinder and \fiberEnder are newly proposed. Specifically, to cope with the challenge that the root location is severely occluded, we formulate root localization as a density map estimation task. Given the predicted density map, a density-based clustering method is further used for finding the roots. For each fiber, the growth starts from the root point and moves step by step until the ending, where each step is defined as an oriented line segment with a constant length according to the predicted orientation field. To determine when to end, a pixel-aligned RNN architecture is designed to form a binary classifier, which outputs stop or not for each growing step. To support the training of all proposed networks, we build the first 3D synthetic eyebrow dataset that contains 400 high-quality eyebrow models manually created by artists. Extensive experiments have demonstrated the effectiveness of the proposed \ems pipeline on a variety of different eyebrow styles and lengths, ranging from short and sparse to long bushy eyebrows.     

\end{abstract}

%
%

\ccsdesc[500]{Computing methodologies~Shape modeling; Neural networks}

%
%

\keywords{eyebrow, deep neural networks, single-view modeling, dataset}

\maketitle

\input{section/1_intro}

\input{figure/fig_pipeline}
\input{section/2_related}
\input{section/3_overview}

\input{section/4_dataset}
\input{section/5_method}
\input{section/6_experiments}
\input{section/7_limitation_and_conclusion}
\input{section/Acknowledgments}

\clearpage
\clearpage
\bibliographystyle{ACM-Reference-Format}
\bibliography{sample-bibliography}

\clearpage
\appendix

\input{Supp/section_Supp/1_implementation}
\input{Supp/section_Supp/2_dataset_length}
\input{Supp/section_Supp/3_density_map_visualization}
\input{Supp/section_Supp/4_more_visualizations_of_EBStore}
\input{Supp/section_Supp/5_more_results_ablation}

\end{document}

%% file: symbols.tex
\newcommand{\dataset}{\emph{EBStore}\xspace}
\newcommand{\rootFinder}{\emph{RootFinder}\xspace}
\newcommand{\oriPredictor}{\emph{OriPredictor}\xspace}
\newcommand{\fiberEnder}{\emph{FiberEnder}\xspace}
\newcommand{\ems}{\emph{EMS}\xspace}

\newcommand{\img}{\mathbf{I}}
\newcommand{\orientmap}{\mathbf{O}}
\newcommand{\eyebrow}{\mathbf{E}}
\newcommand{\head}{\mathbf{H}}
\newcommand{\eyebrowroot}{\mathbf{R}}
\newcommand{\eyebrowdirection}{\mathbf{D}}
\newcommand{\eyebrowender}{\mathbf{L}}
\newcommand{\eyebrowsynther}{Syn}
\newcommand{\eyebrowfiber}{\mathbf{S}}
\newcommand{\densityfunc}{\mathbf{P}}

%% file: figure/fig_teaser.tex
\begin{teaserfigure}
\centering
  \includegraphics[width=1.\textwidth]{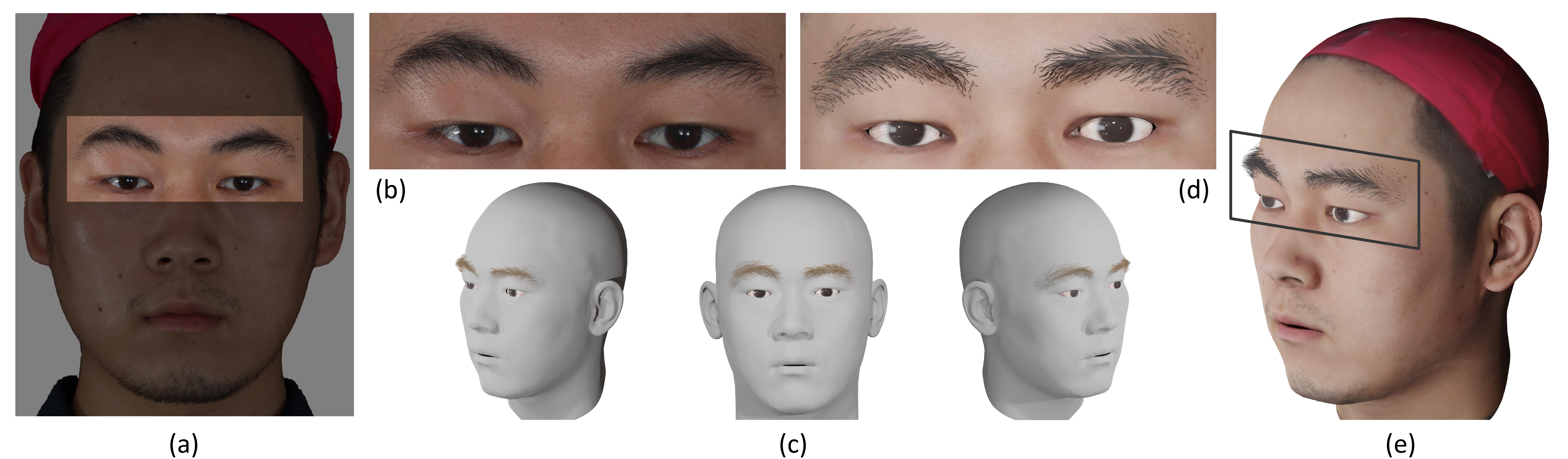}
  \caption{Nowadays, 3D heads can be easily digitalized with high-fidelity geometry and texture, like the setup used in FaceScape. In this work, we proposed \ems, a learning-based approach, which can further reconstruct 3D eyebrows from just the frontal-view image. (a) the input image. (b) the cropped eyebrow image from (a). (c) our reconstructed fiber-level 3D eyebrow model rendered with multiple views. (d) the cropped rendering of (e) for comparison with (b). (e) putting our result on the textured 3D head can further improve the realism of face digitalization.}
  \label{fig:teaser}
\end{teaserfigure}

%% file: section/1_intro.tex
\section{Introduction}

Recently, creating a digital human has become easier along with the development of capturing setups and also deep-learning based approaches. In addition to numerous studies on the digitization of the head and human body \cite{SMPL:2015, FLAME:SiggraphAsia2017, saito2020pifuhd, zheng2021pamir, xiu2022icon, deng2019accurate,  DECA:Siggraph2021,  zielonka2022towards, bao2021high}, there are also many works that pay attention to the details of the facial features due to their vital roles in supporting a high-fidelity face, for example, the modeling of eyes~\cite{berard2014high, berard2016lightweight, li2022eyenerf, bermano2015detailed, wen2017real}, mouth~\cite{garrido2016corrective, zoss2018empirical, zoss2019accurate, dinev2018user, ploumpis20223d} and teeth~\cite{wu2016model, lingchen2020iorthopredictor, velinov2018appearance, zhang2022implicit}. Compared with them, facial hair (including eyebrows, eyelashes, and beards) has not received sufficient attention. Facial hair, particularly eyebrows, also plays a vital role in representing unique personal characteristics. According to ~\cite{sadr2003role}, eyebrows represent a significant influential attribute for facial recognition. To take high-fidelity human digitization one step further, in this work, we focus on the 3D reconstruction of eyebrows. 

To the best of our knowledge, no prior work has been devoted to algorithms specifically tailored for 3D eyebrow modeling. Existing facial hair modeling methods~\cite{beeler2012coupled, winberg2022facial}  primarily focus on recovering the overall facial hair. They adopted multi-view stereo (MVS)~\cite{seitz2006comparison} systems customized to reconstruct 3D facial hair fibers from the appearance cues of multi-view images. These methods rely on specialized hardware and well-controlled settings, making them not widely accessible to the average consumer. To ease the capturing procedure and improve user-friendliness, several works \cite{herrera2010toward, rotger2019single} have been proposed to attempt conducting reconstruction from monocular inputs. However, as the first attempt, their methods are too straightforward and crude, making the results far from natural-looking. Specifically, rule-based methods neglect accurate root point localization and fiber growing direction determination. We argue that the formidable challenges of severe self-occlusion and single-view ambiguity greatly hinder the success of rule-based approaches. Thus, in this work, we propose to use data-driven techniques for single-view 3D eyebrow reconstruction.   



Leveraging data-driven methods is a popular solution frequently employed in the problem of single-view scalp hair reconstruction ~\cite{wu2022neuralhdhair, zheng2023hairstep, zhou2018hairnet}. The incorporation of learned priors from the data can provide information for occluded regions and recover more plausible results. Existing works usually use 3D strands to represent hairs and then the problem becomes the generation of those strand curves. Regarding the extreme complexity of generating a heavy amount of strands, the key idea to simplify this problem is to grow curves from the scalp surface. To be specific, for each strand, they usually considered three questions: where to start; how to grow; when to stop. As all root points of hairs cannot be observed typically, the commonly used way is to pre-sample a fixed number of points (usually 10k) densely on the scalp surface as starting positions. Then, each root point will initiate the growth of a fiber and each fiber will grow step by step, where each step is defined by an oriented line segment with a pre-defined constant length. To guide such growth, an orientation field is required, and inferring an orientation field from the input image thus becomes one of the core problems. Another important issue is determining when to stop the growth of fibers. Existing methods directly used the 3D contour of the hairs for strand cutting, where they typically predicted a 3D occupancy field to define the scalp hair contour mesh.        


For 3D eyebrows, we use fiber (similar in concept to strand with shorter length) as the output format and treat it as a curve growing problem. For the prediction of the orientation field, we directly adopt the method in \cite{wu2022neuralhdhair}. However, we found it is not proper to use existing methods of root localization and strand cutting in hair reconstruction for eyebrow modeling. This is due to the inherent differences between scalp hairs and eyebrows. At first, although the root points of eyebrows may be severely occluded, they can still be partially observed. Thus, an intuitive uniform sampling will produce results that do not match the input image. Second, considering a frontal-view facial image, the fibers of the eyebrow may not only have out-of-plane growth but also in-the-plane growing manner. For in-the-plane growing directions, a 3D contour is usually not sufficient for accurate fiber length control. 


To address the aforementioned challenges, we proposed two novel learning-based methods for root localization and fiber length determination individually. Due to the severe occlusion of the root points, a direct detection solution for root localization tends to fail. We thus introduce the density map as an intermediate representation of the roots. Specifically, we first construct an image-to-image translation network to map the input image into a density map, where each pixel of the density map represents the response of a root existence. Then, a density-based clustering method is employed to extract root locations from the predicted density map. Next, we lift the 2D root points to 3D positions. The above density map prediction network combined with the clustering and lifting method is termed as \rootFinder. After all root points are confirmed, each root will grow a fiber, which will then grow step by step. Similar to the method in hair reconstruction, each growing step follows an orientation field which is predicted based on a pixel-aligned implicit learning framework~\cite{wu2022neuralhdhair}. The prediction method of the orientation field is termed as \oriPredictor. For each growing fiber, we observed that determining when to stop depends not only on its location but also on the image appearances along its growth trajectory. Based on this assumption, we design a pixel-aligned stacked RNN~\cite{giles1994dynamic} architecture to form a classifier that outputs a binary status for each growing step: stop or continue. To be specific, the network is conditioned on both a positional encoding of the sequential 3D points and an accumulated pixel-aligned image feature along the historical growing path. We term this as \fiberEnder.      


To summarize, we introduce \ems in this work, which is the first deep learning-based framework to reconstruct 3D eyebrow models from a single portrait image. Our \ems consists of three modules: \rootFinder, \oriPredictor and \fiberEnder. To support both the training and evaluation of all three modules, we also contribute a dataset with well-annotated synthetic fiber-level eyebrows of 400 people in FaceScape~\cite{yang2020facescape}, which are manually created by artists. Extensive experiments have demonstrated the superiority of our method over existing rule-based methods and the effectiveness of all our module designs. The results of in-the-wild testing also verify the generalization ability of our method, even when it is trained using synthetic data only.

The main contributions of this work are listed as follows:
\begin{itemize}
\item We propose the first learning-based framework for 3D eyebrow modeling, capable of reconstructing fiber-level eyebrow geometry from only a single portrait image. Our method achieves state-of-the-art performance.
\item Considering the differences among eyebrows and hairs, we propose a new method for root point localization which leverages a density map as an intermediate representation. 


\item We are the first to formulate fiber length determination of eyebrows as a classification problem and design a novel pixel-aligned RNN architecture for solving it. 


\item We contribute the first high-quality 3D synthetic eyebrow dataset \dataset, which will be released to benefit the research in this area.
\end{itemize}


%% file: figure/fig_pipeline.tex
\begin{figure*}[t]
\centering
\includegraphics[width=1.0\linewidth]{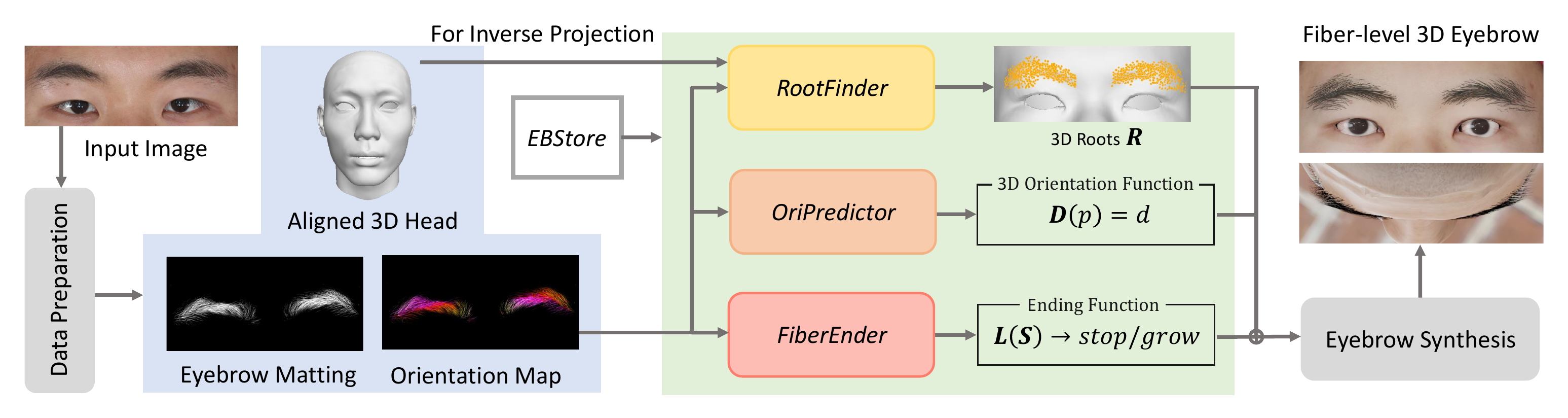}
\caption{The pipeline of eyebrow modeling from single-view images.
Given a single-view image, our method reconstructs a fiber-level 3D eyebrow. 
We first prepare the intermediate data, then feed them into \rootFinder, \oriPredictor and \fiberEnder to generate 3D eyebrow roots $\eyebrowroot$, 3D orientation function $\eyebrowdirection$ and ending function $\eyebrowender$.
Following the eyebrow synthesis, we finally obtain the fiber-level eyebrow model.}
\label{fig:overview}
\end{figure*}

%% file: section/2_related.tex
\section{Related work}
This section provides a comprehensive review of existing works closely related to our work. \\

\textbf{Scalp Hair Capturing.} Scalp hair modeling has been extensively explored in computer graphics over the past few decades. \citeauthor{grabli2002image}~[\citeyear{grabli2002image}] utilized hair scattering profiles to recover 3D hair orientation from image sequences captured under moving light sources. Follow-up works~\cite{ luo2012multi, paris2004capture, Wei2005modeling, paris2008hair} estimate a dense vector field from different views or under different illuminations with 3D geometry constraints, e.g., visual hull, structure light, or epipolar constraints. To improve the hair capturing quality, various sensor modalities have also been explored, such as depth-of-focus devices \cite{jakob2009capturing}, thermal imaging cameras~\cite{herrera2012lighting}, or RGB-D sensors~\cite{hu2014capturing}. \citeauthor{luo2013structure} [\citeyear{luo2013structure}] and \citeauthor{hu2014robust} [\citeyear{hu2014robust}] used shape primitives like ribbons and wisps to fit a fused point cloud generated by MVS for the acquisition of complete hair strands that connect to the scalp. Additionally, \citeauthor{xu2014dynamic}~[\citeyear{xu2014dynamic}] traced a motion path for each hair pixel for dynamic hair temporal coherence. \citeauthor{nam2019strand}~[\citeyear{nam2019strand}] introduced a slanted line-based MVS framework with a novel cost function. \citeauthor{sun2021human}~[\citeyear{sun2021human}] leveraged a per-pixel lightcode to boost the multi-view consistency of 3D hair segments and estimate hair reflectance for realistic rendering. \citeauthor{rosu2022neural}~[\citeyear{rosu2022neural}] attempted to recover hair strand geometry and appearance by differentiable rendering with multi-view inputs for the first time and explored a novel representation that parameterizes a hairstyle using a latent texture map on the scalp. To simplify the data acquisition, several works investigate modeling hair from sparse views~\cite{zhang2017data, zhang2018modeling} and selfie videos~\cite{liang2018video}. Recently, \citeauthor{kuang2022deepmvshair}~[\citeyear{kuang2022deepmvshair}] introduced a deep learning-based framework for multi-view strand hair modeling that enables the reconstruction of view-consistent hair geometry using only sparse views. However, the complex hardware setup or long processing cycles make these techniques inaccessible to the average user. \\

\textbf{Facial Hair Capturing.} Compared to the extensive research on digital modeling of human scalp hair, facial hair capturing has been a relatively neglected research area. Pioneer work by~\cite{herrera2010toward} attempts to model static facial hair using statistical characteristics of texture images, growing 3D hairs through particle shooting techniques, and modeling facial hair geometry roughly. To generate connected piece-wise 3D facial hair primitives on a static face, \citeauthor{beeler2012coupled} [\citeyear{beeler2012coupled}] proposed a passive, multi-camera system with uniform lighting, which incorporates 2D hair detection and 3D facial hair growing algorithms. While capturing individual fibers of facial hair is available, their approach involves considerable post-processing and refinement steps to generate plausible hair segments. Furthermore, their system also requires additional close-up photographs of the facial hair regions. Contemporary work by~\cite{fyffe2012high}  introduces a novel photo-consistency cost function to help identify areas containing facial hair and triangulate 2D line segments to a set of 3D oriented hair particles from multiple views. Additionally, \citeauthor{legendre2017modeling}~[\citeyear{legendre2017modeling}] employed asperity scattering and hair fiber geometric parameter fitting techniques along vellus hairs' backlit silhouette to model barely noticeable, fine texture vellus hair. Inspired by~\cite{beeler2012coupled}, \citeauthor{rotger2019single}~[\citeyear{rotger2019single}] extended the task formulation with just a single-view image as input, but with limited fidelity and realism due to the representation of 3D facial hair fibers with solely a simple hair growing parametric model. Recently, \citeauthor{winberg2022facial}~[\citeyear{winberg2022facial}] improved upon~\cite{beeler2012coupled} to integrate dense facial hair tracking and underlying skin deformation capabilities. Although their method obtains impressive results in capturing beard hair from dynamic sequences, it struggles to capture eyebrow geometry with high-frequency details. The primary cause of this limitation is the inherent self-occlusion problem in eyebrows, which is exacerbated for thicker, bushier eyebrows. Consequently, the precision of identifying hair follicle locations and multi-view hair fiber matching is diminished even with the input of multiple front-view images. These factors greatly affect the fidelity and realism of the reconstructed eyebrows and highlight the need for more specialized techniques to address the eyebrow modeling complexities.\\

\textbf{Single-View Scalp Hair Modeling.} In contrast to multi-view scalp hair reconstruction techniques, single-view hair modeling methods~\cite{chai2012single, chai2013dynamic} demonstrate notable advantages in generality and efficiency. Later works by~\cite{hu2015single, chai2016autohair} provide synthetic 3D hair databases and generate satisfactory results from a single portrait image based on data-driven techniques. However, the fidelity of their hair reconstruction results is highly dependent on the quality and diversity of the hair dataset. In particular, the retrieved hair model is prone to fail if a 3D hair model with an identifiable likeness is not available in the database. Recently, the significant advancements achieved in deep neural networks across various fields have prompted the adoption of learning-based algorithms for scalp hair modeling. Early work by~\cite{zhou2018hairnet} exploits an encoder-decoder network to infer 3D hair strands directly from a 2D orientation map of the segmented hair region. \citeauthor{saito20183d} [\citeyear{saito20183d}] trained a variational autoencoder to represent 3D hairstyles as a latent space of hair volume and utilized the latent code to synthesize 3D hair from a single-view image. Later work by~\cite{zhang2019hair} introduces a generative adversarial network with 3D convolutional layers to recover the 3D orientation field of hair strands. Furthermore, \citeauthor{yang2019dynamic}~[\citeyear{yang2019dynamic}] investigated dynamic hair modeling from monocular video streams and inferred 3D spatial and temporal features of moving hairs via neural networks. While the aforementioned methods are capable of capturing the overall shape of hairs and predominant strand orientation, their explicit hair representation (e.g. orientation voxels) tends to reconstruct over-smoothed 3D hair geometry. To overcome this limitation, concurrent works by~\cite{wu2022neuralhdhair, zheng2023hairstep} leverage implicit representation and pixel-aligned features to represent intricate hair geometry. Despite a large body of research devoted to learning-based scalp hair modeling, the application of these methods to eyebrow reconstruction is not adequate due to the notable differences in the root distributions and growing rules between scalp hair and eyebrows.


%% file: section/3_overview.tex
\section{Overview}

This section contains a system overview for the single-view 3D reconstruction of a fiber-level eyebrow model.
We first give the definition of this problem, then briefly introduce our pipeline.

\subsection{Problem Definition}

Given a single input image $\img$, we need to recover the corresponding fiber-level 3D eyebrow model $\eyebrow$ growing on the target 3D head $\head$. 

\textbf{3D eyebrow representation.}
A fiber-level 3D eyebrow model $\eyebrow$ is actually a set of 3D curves growing from the roots to the tips. 
Following state-of-the-art methods of scalp hair modeling~\cite{wu2022neuralhdhair,zheng2023hairstep}, we model $\eyebrow$ via an implicit orientation field. 
Once we obtain the eyebrow roots and lengths, fiber curves can be easily grown step-by-step from this field.
Thus, we define the fiber-level 3D eyebrow model as 
\begin{equation}
\label{eqn:01}
\begin{aligned}
\eyebrow=\eyebrowsynther(\eyebrowroot, \eyebrowdirection, \eyebrowender),
\end{aligned}
\end{equation}
where $\eyebrowsynther(\cdot)$ stands for the algorithm of eyebrow synthesis, i.e. the growing procedure. 
$\eyebrowroot$, $\eyebrowdirection$ and $\eyebrowender$ represent the eyebrow roots, implicit 3D orientation function and the ending function, respectively. 
Specifically, the eyebrow roots $\eyebrowroot=\{r_i,0\le i < n\}$ is a 3D point cloud attached to the surface of the brow bone region on the target head $\head$, with a dimension of $n\times3$. 
The implicit orientation function $\eyebrowdirection$ maps a 3D query point $p$ to a unit growing vector $d$, which can be formulated as $d = \eyebrowdirection(p)$. 
The ending function $\eyebrowender$ describes whether a growing eyebrow fiber $\eyebrowfiber$ should be stopped:
\begin{equation}
\label{eqn:02}
\begin{aligned}
\eyebrowender(\eyebrowfiber)=\begin{cases}0, & \text { if } \eyebrowfiber \text { should be stopped } \\ 1, & \text { otherwise }\end{cases}.
\end{aligned}
\end{equation}

%

\subsection{Pipeline}

\cref{fig:overview} shows the pipeline of our learning-based framework (\ems) for single-view eyebrow modeling. Our method takes a single portrait image as input and outputs a fiber-level 3D eyebrow model with fine details. Given the lack of a 3D eyebrow dataset, we create a high-quality synthetic 3D eyebrow dataset named \dataset (\cref{sec:dataset}), to facilitate the data-driven approach. 
Based on \dataset, \ems achieves state-of-the-art performance on single-view 3D eyebrow modeling.

Starting from a cropped portrait of the eyebrow region, we first prepare the intermediate data, including the 3D head, eyebrow matting, and the orientation map (\cref{subsec:preparation}). Following the definition in~\cref{eqn:01}, we break down 3D eyebrow modeling into three steps, i.e. localizing fiber root (\cref{subsec:rootfinder}), predicting 3D orientation (\cref{subsec:oripredictor}) and determining fiber length (\cref{subsec:fiberender}). 
After the eyebrow synthesis (\cref{subsec:fibersyn}), we finally obtain a fiber-level 3D eyebrow model.
It should be noted that, as there is a non-negligible domain gap between real images and synthetic images, we follow~\cite{zhou2018hairnet,kuang2022deepmvshair,wu2022neuralhdhair} to adopt the 2D orientation map as the intermediate representation rather than input the image directly to the sub-modules.

%% file: section/4_dataset.tex
\section{Dataset Construction}
\label{sec:dataset}

\input{figure/fig_EBStore_dataset}
Unlike the existing facial hair datasets~\cite{eyebrows_dataset, wang2022effective, xiao2021eyelashnet} with only 2D detection and alpha matting annotations, we construct the first 3D fiber-level synthetic eyebrow dataset spanning a wide range of eyebrow geometries and different local details based on FaceScape~\cite{yang2020facescape}. 
FaceScape obtained data from a dense multi-view stereo with DSLR cameras, which provides high-quality 3D face models. 
These face models are registered to the template head mesh with uniform topology and located within a unified space, which facilitates the rough alignment of 3D eyebrows attached to various brow bone geometries.
Besides, 68 DSLR cameras of the multi-view system are able to capture 4K images, ensuring clear visibility of the eyebrows in frontal views.
We carefully selected 197 males and 203 females from the large-scale FaceScape dataset to ensure gender balance in building the \dataset dataset.
The subjects in \dataset range from 16 to 70 years old with diverse eyebrow styles. 
These 400 3D eyebrow models are manually created by artists using particle edit mode in Blender~\cite{blender}, guided by both the high-resolution texture map of the eyebrow region and a front-facing image of the individual. 
In general, a skillful artist typically needs to take around 1-2 days to create a high-quality fiber-level eyebrow model that closely matches the given image.
Each eyebrow fiber in \dataset is comprised of 20 connected 3D points for further processing. 
\cref{fig:dataset} shows a gallery of our data.

%% file: figure/fig_EBStore_dataset.tex
\begin{figure*}[t]
\centering
\includegraphics[width=1.0\textwidth]{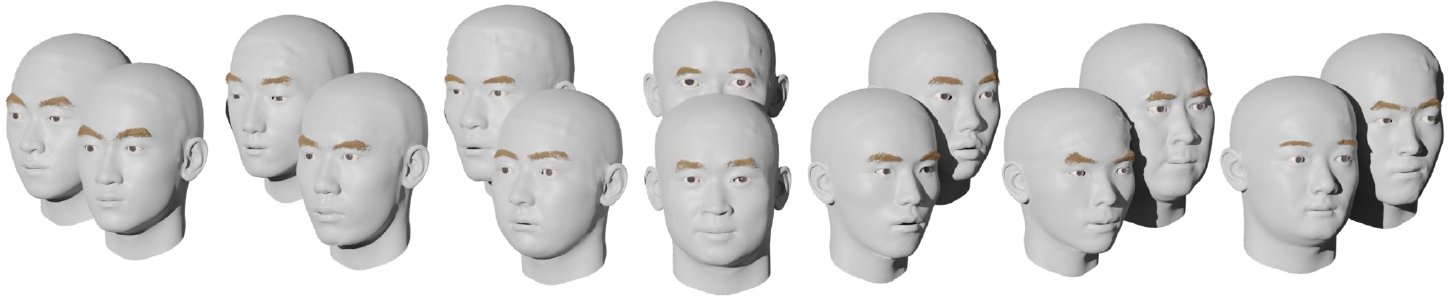}
    \vspace{3mm}   
\caption{\dataset dataset gallery with diverse 3D eyebrow models.} 
\label{fig:dataset}
\end{figure*}

%% file: section/5_method.tex
\section{Methodology}
\label{sec:method}
In this section, we describe the detailed designs of our approach.

\subsection{Preparation}
\label{subsec:preparation}
The initial stage of the preparation for our system is to detect eyebrow regions from single-view portrait images. We use the 3DDFA method ~\cite{guo2020towards} to detect the face region of the input image and then feed the face image to the pre-trained face parsing model on CelebAMask-HQ~\cite{lee2020maskgan}. Then we crop the eyebrow region from the face parsing results and adopt the eyebrow matting method~\cite{wang2022effective} to get the mask of eyebrow fibers. Next, we compute the real orientation map $\orientmap$ for the eyebrow fiber map with the mask using the method of~\cite{luo2012multi},  which uses a bank of rotated filters to detect the dominant orientation at each pixel. The orientation map $\orientmap$ is then enhanced with 3 passes of iterative refinement for a better signal-to-noise ratio proposed by \cite{chai2012single}. As for acquiring the human head $\head$, we get a detailed 3D human face from a single-view 3D face prediction method~\cite{yang2020facescape}. By utilizing the mapping between the 2D image and 3D face mesh, we can derive a pseudo-camera pose for eyebrow projection. However, it should be noted that the predicted 3D face mesh may not be aligned with our training head space. To solve this, we align the predicted face mesh with the average head of FaceScape by optimizing its translation, rotation and scale.   

\subsection{Localization of Fiber Root}
\label{subsec:rootfinder}
Given the eyebrow orientation map $\orientmap$ as input, we aim to find the 3D roots $\eyebrowroot$ of the eyebrow fibers. As previously discussed, directly detecting eyebrow roots encounters a tough challenge due to severe self-occlusion from visual perception.
To address this issue, we proposed a density-based eyebrow root localization method, named \rootFinder.

In our approach, the root distribution estimation of the eyebrow is viewed as the problem of density map prediction, which involves the modeling of the spatial distribution of root points as a continuous function over the image domain. The high-level idea of our approach is quite straightforward: given an eyebrow orientation map $\orientmap$, our goal is to recover a density function $\densityfunc$ that approximates the real distribution of each fiber root pixel in this image. Our notion of a density function $\densityfunc$ can be interpreted as the response of a given pixel representing a root point. To prepare the training data, we employ the pre-processed orientation maps as input and utilize their corresponding cameras to project 3D ground-truth root points to the 2D image coordinates. Motivated by the previous crowd counting methods~\cite{liu2019context, ranjan2021learning}, we generate the ground-truth density map,  $P^* \in \mathbb{R}^{H \times W}$,  from the 2D root coordinates by using a Gaussian kernel with adaptive window size. Inspired by the encoder-decoder crowd counting framework~\cite{jiang2019crowd, yan2019perspective}, we find that simply using a U-Net based architecture~\cite{ronneberger2015u} can already achieve satisfactory results. Following the standard settings, we use a pixel-wise MSE loss against $P^*$ , which is formulated as:
\begin{equation}
    L_{den} = \dfrac{1}{N} \sum_{i =1} ^{N}{||P(O_i) - P_{i}^{*}||^2},
\end{equation}
where $P(O_i)$ is the predicted density map. $O_i$ is the $i^{th}$ input of total $N$ orientation maps, and $P_{i}^*$ is the ground-truth density map.

\input{figure/fig_rootfinder}
\input{figure/fig_fiberender}
Since predicting the distribution of eyebrow roots involves a rough estimation of their coordinates, the output density map at the patch level cannot be directly used for accurate pixel-level root localization. Thus we employ DBSCAN~\cite{ester1996density}, a density-based clustering algorithm, to compute the number of valid fiber root clusters. Once we determine the number of clusters using DBSCAN, we apply the K-Means~\cite{hartigan1979k} clustering method to calculate the cluster center, which corresponds to the 2D root points. Next, our goal is to derive the 3D positions of the acquired 2D root coordinates. However, directly lifting the 2D root points to 3D space is an ill-posed problem since inverse projection leads to depth ambiguity. To overcome this issue, we first sample dense 3D points on the brow bone surface region of the face mesh and project them onto the image. For each extracted 2D root, we find its nearest projected 2D sampling points and take the corresponding 3D sampling point as the 3D root $r_i$. Relative maps mentioned in this subsection are illustrated in~\cref{fig:rootfinder}.

\subsection{Prediction of 3D Orientation}
\label{subsec:oripredictor}
To recover the growing direction of fiber $\eyebrowfiber$, $\oriPredictor$ is constructed to predict a unit vector for every point on $\eyebrowfiber$ under the guidance of the given image $\img$. In this task, the pre-processed 2D orientation map $\orientmap$ from $\img$ is used to calculate the feature map $F$, and the 3D point $p$ is projected to 2D image coordinate by camera projection $\pi(\cdot)$ to exact pixel-aligned local image feature $F(\pi(p))$. In addition, to represent high-frequency variation in eyebrow fiber geometry, positional encoding $\Phi(p)$ is added to transform coordinates to higher dimensional information~\cite{mildenhall2021nerf}. Then we can define the implicit orientation function $\eyebrowdirection$ for arbitrary point $p$ in 3D space as
\begin{equation}
    \mathbf{D}(F(\pi(p)),\Phi(p)) = d:d \in \mathbb{R}^3 .
\end{equation}

In our implementation, we apply an Hourglass filter to obtain feature maps from input orientation maps. Positional encoding will be concatenated with the image feature as a joint local feature. MLPs are designed as the decoder to output a unit 3D vector $d$ and $L_1$ loss is chosen to measure the error between the predicted direction $d$ and corresponding ground-truth direction $d^*$ during training.

\subsection{Determination of Fiber Length}
\label{subsec:fiberender}
Theoretically, a growing eyebrow fiber $\eyebrowfiber$ with $q$ points can be discretized as a 3D points sequence $\{p_0,p_1,...,p_{q-1}\}$. \fiberEnder is designed to determine whether $\eyebrowfiber$ is supposed to be stopped by emitting an ending signal, which can be formulated into a binary classification problem. 

It is trivial to think that one can tell the ending point of an eyebrow fiber on a close-up photo based on its surrounding pixels, so the local image feature should be considered. We also leverage 3D information by encoding fiber point coordinates into the local feature list since images may not be able to exhibit useful details in the region of self-occlusion. Hence, the concatenated growing feature $G$ for a single 3D point $p$ can be given by
\begin{equation}
    G(p) = \{F(\pi(p)),\Phi(p)\},
\end{equation}
where $\pi(\cdot)$ is the camera projection. $F(\pi(p))$ and $\Phi(p)$ are the image feature and positional feature for $p$, respectively.

However, noticed that features of points along the same fiber have a strong correlation, we cannot treat a node as an isolated identity, since the prediction of whether $\eyebrowfiber$ will be stopped at $p_{q-1}$ is highly dependent on the features of its predecessor. Therefore, the ending function $\eyebrowender$ for $\eyebrowfiber$ can be defined as
\begin{eqnarray}    \label{eq}
\eyebrowender(\eyebrowfiber) &=& \eyebrowender(\{p_0,p_1,...,p_{q-1}\})  \nonumber   \\
~&=& C(G(p_0),G(p_1),...,G(p_{q-1})) \nonumber    \\
~&=& \begin{cases}
\begin{aligned}
0,\quad& {\text{if\:} \eyebrowfiber\: \text{should be stopped at\:} p_{q-1}}\\
1,\quad& {\text{otherwise}}
\end{aligned}
\end{cases},
\end{eqnarray}
where $C$ is a composite function taking the indefinite number of point feature $G$ as ordinal inputs. 

In practice, consistent with the prediction of the  3D orientation field, the Hourglass filter is employed to exact 2D image local features from input orientation maps. For organizing a point feature series with an unfixed length into a latent vector, we borrow knowledge from ~\cite{schuster1997bidirectional} and introduce a stacked RNN based encoder. The iteration begins with the initial zero hidden code $h_0$ and it will be passed to a GRU cell~\cite{cho2014learning} together with a new growing feature $G(p_i),0\le i<q$ for total $q$ times. Then the final hidden code $h_q$ is passed to a MLP classifier to output a probability that $\eyebrowfiber$ will continue growing. It is worth mentioning that the direction of fiber growth is not reversible, thus we use a monodirectional stacked RNN rather than a bidirectional one. The whole structure of \fiberEnder is illustrated in~\cref{fig:fiberender}.

Based on \dataset, we annotate all fibers and their sub-sequences with binary labels: the full sequence with the tip point is labeled to $l=0$ as the negative sample $S^-$, indicating the fiber will be stopped; the sub-sequences are labeled to $l=1$ as the positive sample $S^+$, suggesting the fiber will continue growing. 
In the training phase, we construct a data pair of $\{S^-, S^+\}$ for each fiber in every epoch, where the positive sample is randomly picked from all its sub-sequences. Binary Cross Entropy (BCE) is chosen to calculate the loss between predicted normalized probability $C(S)$ and labels:

\begin{equation}
    L_{BCE} = -(l\cdot log(C(S))+(1-l)\cdot log(1-C(S))).
\end{equation}

\subsection{Synthesis of Fiber-level Eyebrow}
\label{subsec:fibersyn}
After $\{\eyebrowroot, \eyebrowdirection, \eyebrowender\}$ has been well-defined, the subsection is aimed to illustrate how to generate a fiber-level eyebrow $\eyebrow$. The whole growing process can be generalized as $\eyebrowsynther(\eyebrowroot, \eyebrowdirection, \eyebrowender)$: Given $n$ detected root points $\eyebrowroot=\{r_i,0\le i < n\}$, $n$ fibers will be generated by point-wise querying from $r_i$. For the $j^{th}$ node $p_{i,j}$ on fiber $\eyebrowfiber_i$, the unit growing direction $\eyebrowdirection(p_{i,j})=d_{i,j} \in \mathbb{R}^3$ and a ending signal $\eyebrowender(\eyebrowfiber_i)=\eyebrowender(\{p_{i,0},p_{i,1},...,p_{i,j}\})=l_{i,j} \in \{0,1\}$. $\eyebrowfiber_i$ will keep growing as the direction of $d_{i,j}$ with a fixed step $\bar{s}$ if $l_{i,j} = 1$. Otherwise, $\eyebrowfiber_i$ will be stopped at point $p_{i,j}$. The above synthesis procedure can be presented in~\cref{fig:fiber}. The step-wise iteration can also be summarized into an algorithm in~\cref{tab:eyebrow_synthesis} (In practice, it can be conducted in batch for all fibers of an eyebrow to achieve computational efficiency).

In addition, considering that the inevitable noises of filter-based orientation maps from real images cause the vibration of the 3D orientation field prediction, we also modify the predicted growing direction $d$ to ensure smoothness and alleviate abnormal warping of fibers. Following ~\cite{shen2020deepsketchhair}, the growing direction is set to the mean value of the current and the last node if their angle difference is larger than a particular threshold $\theta$. However, the eyebrow fibers tend to exhibit less drastic angle variation between two adjacent points compared with hair strands, thus the given threshold in hair modeling is no more suitable for eyebrow modeling. Based on our experimental findings, we set $\theta$ to be $30^\circ$.  

\input{table/table_growing_algorithm}
\input{figure/fig_fiber_synthesis}

%% file: figure/fig_rootfinder.tex
\begin{figure}
\centering
\includegraphics[width=1.0\linewidth]{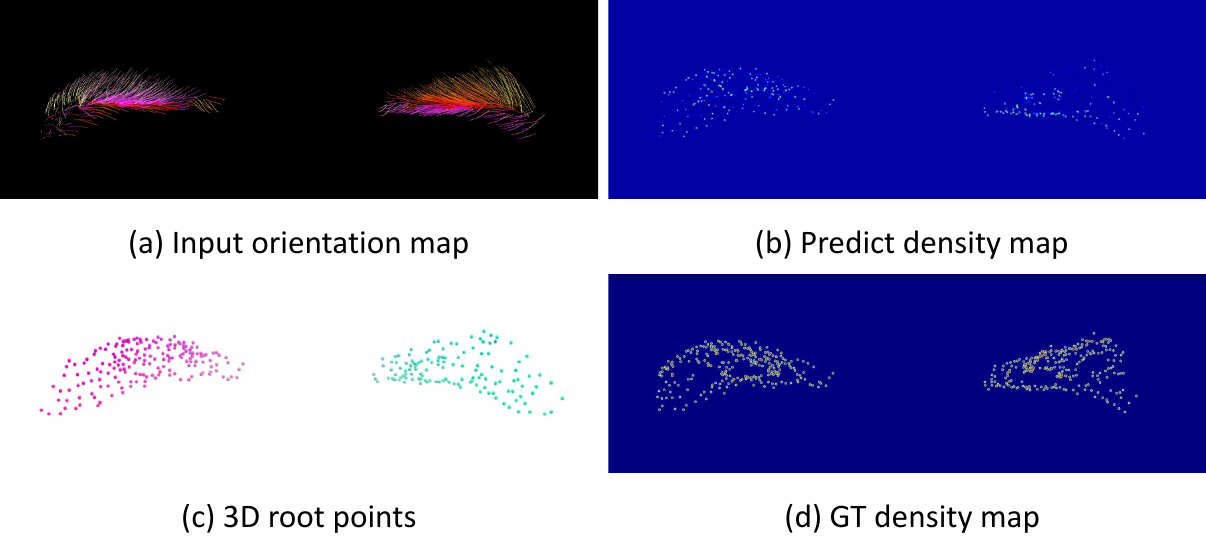}
\caption{Localization of the 3D root points by \rootFinder. (a) input orientation map. (b) predict density map. (c) 3D root point cloud by clustering and lifting. (d) ground-truth density map.}
\label{fig:rootfinder}
\end{figure}

%% file: figure/fig_fiberender.tex
\begin{figure*}
\centering
\includegraphics[width=1.0\linewidth]{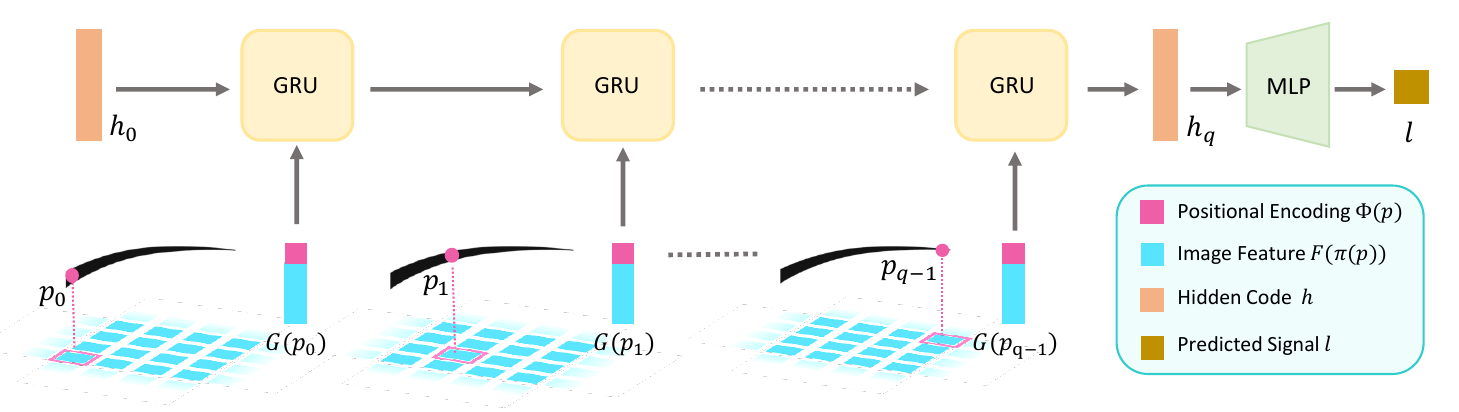}
\caption{The architecture of the proposed pixel-aligned stacked RNN-based \fiberEnder. Radius change along a fiber has no particular meaning, only for intuitive illustration.}
\label{fig:fiberender}
\end{figure*}

%% file: table/table_growing_algorithm.tex
\begin{table}
\small
\begin{tabular}{l}
\hline \textbf{ALGORITHM:} Fiber-level Eyebrow Synthesis \\
\hline 
\textbf{Input:} root points $\eyebrowroot$, 3D orientation function $\eyebrowdirection$, ending function $\eyebrowender$ \\
\textbf{Output:} fiber-level eyebrow $\eyebrow$ \\

\textbf{for} $i = 0 : n$ \textbf{do}\\
\qquad build the empty point list $\eyebrowfiber_i$ of the $i^{th}$ fiber \\
\qquad $p_{i,0} = r_i$ \\
\qquad $\eyebrowfiber_i = \{\eyebrowfiber_i,p_{i,0}\}$ \\
\qquad \textbf{while} $j = 0 : \infty$ \textbf{do}\\
\qquad \qquad $d_{i,j} = \eyebrowdirection(p_{i,j})$ \\
\qquad \qquad $l_{i,j} = \eyebrowender(\eyebrowfiber_i)$ \\
\qquad \qquad \textbf{if} $l_{i,j} = 0$: \textbf{break}\\
\qquad \qquad $p_{i,{j+1}} = p_{i,j} + \bar{s}d_{i,j}$ \\
\qquad \qquad $\eyebrowfiber_i = \{\eyebrowfiber_i,p_{i,{j+1}}\}$ \\
\qquad \textbf{end}\\
\textbf{end}\\
$\eyebrow = \{\eyebrowfiber_0,\eyebrowfiber_1,...,\eyebrowfiber_{n-1}\}$ \\

\hline
\end{tabular}
    \vspace{3mm}   
\caption{Growing algorithm of the fiber-level eyebrow.}
 \label{tab:eyebrow_synthesis}
\end{table}

%% file: figure/fig_fiber_synthesis.tex
\begin{figure}
\centering
\includegraphics[width=0.9\linewidth]{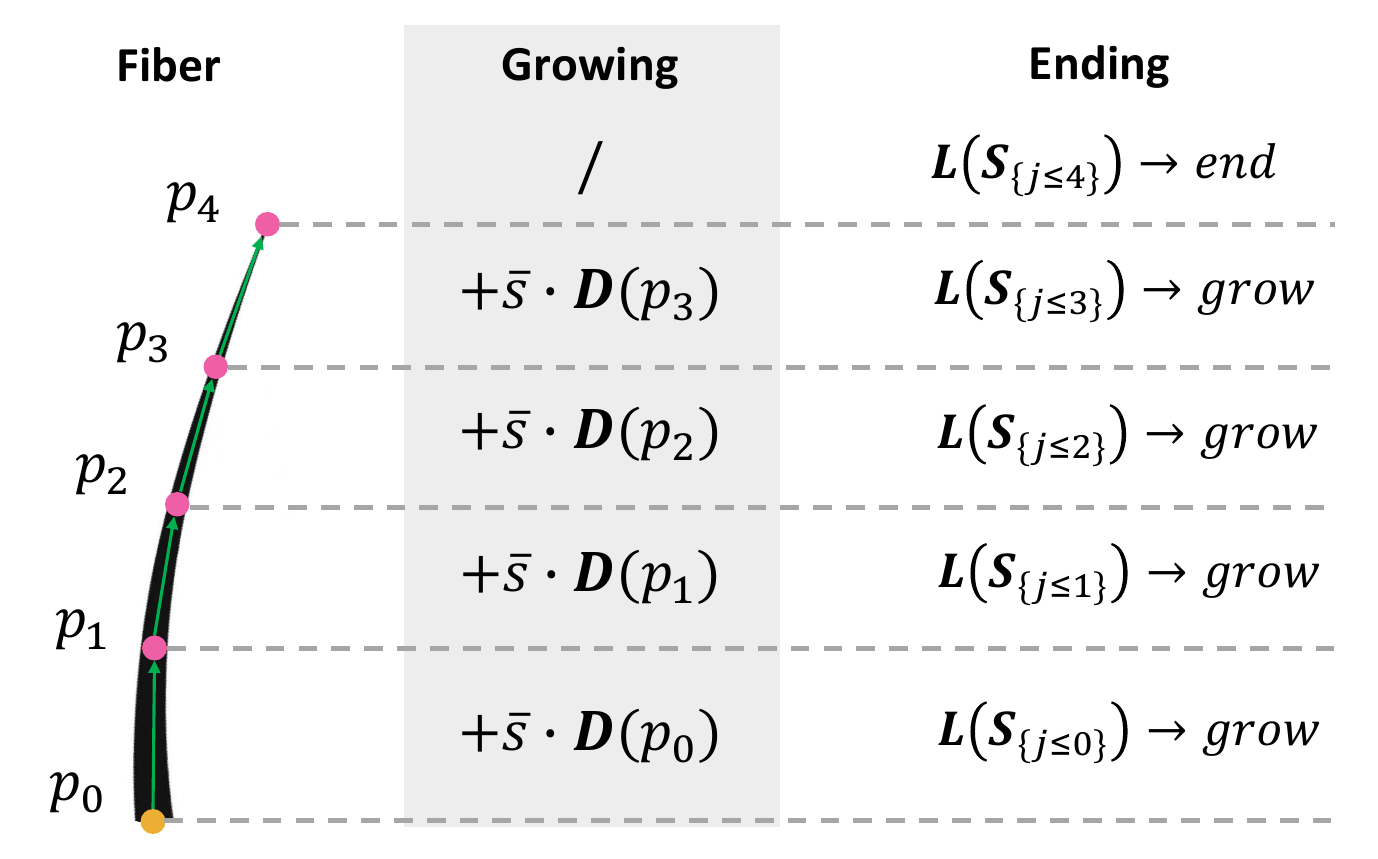}
\caption{Synthesis of an eyebrow fiber. 
Starting from a root point in $\eyebrowroot$, an eyebrow fiber is grown by querying the 3D orientation function $\eyebrowdirection(p)$  and the ending function $\eyebrowender(S)$ step-by-step. Radius change along a fiber has no particular meaning, only for intuitive illustration.}
\label{fig:fiber}
\end{figure}

%% file: section/6_experiments.tex
\section{Experiments}
\label{sec:experiment}
\subsection{Dataset}
Our experiments are based on 400 synthetic eyebrow models in \dataset. For each 3D model, we employ flipping w.r.t the reflection symmetry plane of the head model to augment the 3D eyebrow dataset and randomly select 3 views by moving the camera position in spherical coordinates with a fixed radius, azimuthal angle in $\pm10^\circ$ and polar angle in $\pm15^\circ$. The camera ray is ensured to pass through the center point of the 3D eyebrow model. 
For each view of an eyebrow model, we render an orientation map.
In total, we have 2202 images for training and 198 images for testing. Based on the observation that hair fibers may grow between the left and right eyebrow in some eyebrow types, we process the left and right parts simultaneously in all experiments instead of separating them for individual processing.

\subsection{Metrics}
In this subsection, we introduce fair metrics for the evaluation of 3D eyebrow modeling. We design several metrics to evaluate root localization and fiber length determination, respectively. Also, we use 3D metrics to assess the overall reconstruction performance of the whole system.

\textbf{Root localization metrics.}
We first compute the nearest density error ($NDE$) and a density-aware chamfer distance ($DCD$) to evaluate the localization of 3D eyebrow roots. 
The nearest density error is based on the density $den(r,\eyebrowroot)$ of root $r$ against the root point cloud $\eyebrowroot$, which is calculated by counting the number of neighbors within a radius $\varphi$. Then the nearest density errors from predicted roots $\eyebrowroot^{'}$ to the ground-truth $\eyebrowroot^{*}$ can be defined as 
\begin{equation}
    NDE^{'} = \frac{1}{\left|\eyebrowroot^{'}\right|} \sum_{x \in \eyebrowroot^{'}} \left|den(x, \eyebrowroot^{'})-den(x^*, \eyebrowroot^{*})\right|,
\end{equation}
where $x^* \in R^{*}$ is the nearest root of $x \in R^{'}$.
Based on the density errors, we also compute a density-aware chamfer distance from predicted roots $\eyebrowroot^{'}$ to the ground-truth $\eyebrowroot^{*}$
\begin{equation}
    DCD^{'} = \frac{1}{\left|\eyebrowroot^{'}\right|} \sum_{x \in \eyebrowroot^{'}} \left|den(x, \eyebrowroot^{'})-den(x^*, \eyebrowroot^{*}) + 1\right| \cdot \|x-x^*\|_2.
\end{equation}
The reversed $NDE^{*}$ and $DCD^{*}$ can be calculated in the same way. 
We take their arithmetic means $NDE = (NDE^{'}+NDE^{*})/2$ and $DCD = (DCD^{'}+DCD^{*})/2$ as the bi-directional metrics to evaluate the root localization.

\textbf{Fiber length determination metrics.}
We define the Mean Length Error ($MLE$) for evaluating the average loss per fiber for $k$ synthetic 3D eyebrow, which can be calculated as
\begin{equation}
    MLE = \frac{\sum_{i=1}^k\sum_{n=1}^{n_i}\bar{s}\:|l_{i,j}-l_{i,j}^*|}{\sum_{i=1}^kn_i},
\end{equation}
where $n_i$ is the total number of fibers of the $i^{th}$ $(1\:{\le}\:i\:{\le}\:k)$ eyebrow, $l_{i,j}$ represents the predicted length labels for its $j^{th}$ $(1\:{\le}\:j\:{\le}\:n_i)$ fiber, and $l_{i,j}^*$ stands for its corresponding ground-truth label. $\bar{s}$ is the pre-determined fixed step.

\textbf{3D fiber-level eyebrow reconstruction metrics.}
For the 3D metrics, we use the IoU of meshes converted from reconstructed eyebrows $\eyebrow^{'}$ and ground-truth eyebrows $\eyebrow^{*}$ via the method in~\cite{zhu2005animating}, which is designed to evaluate the overall shape, i.e. the silhouette of the reconstructed eyebrow model.
To further compare the inside geometry of eyebrow growing, we evaluate the fiber-level $L_2$ distance of 3D orientation, which is shorted as $FDO$. For the $FDO^{'}$ from $\eyebrow^{'}$ to $\eyebrow^{*}$, we first find the nearest fiber $\eyebrowfiber^{*} \in \eyebrow^{*}$ for each fiber $\eyebrowfiber \in \eyebrow^{'}$ according to the distance of their roots, then compute $FDO^{'}$ as 
\begin{equation}
    FDO^{'} = \frac{1}{\left|\eyebrow^{'}\right|} \sum_{\eyebrowfiber \in \eyebrow^{'}} \sum_{p \in \eyebrowfiber, q \in \eyebrowfiber^{*}} \|\eyebrowdirection(p_i) - \eyebrowdirection(q_i)\|_2,
\end{equation}
where $\eyebrowdirection(p_i)$ and $\eyebrowdirection(q_i)$ are the unit growing directions of the $i^{th}$ points $pi$ and $qi$ along $\eyebrowfiber$ and $\eyebrowfiber^{*}$, respectively. We use the bi-directional error $FDO = (FDO^{'}+FDO^{*})/2$ as the metric.  

\subsection{Evaluations}
This subsection is to evaluate our key modules \rootFinder, \oriPredictor and \fiberEnder on the synthetic test set by both quantitative and qualitative results.

\textbf{Evaluation of eyebrow roots.}
Firstly, we analyze the performances of \rootFinder by calculating $DCD$ and $NDE$. 
We compare our \rootFinder with an intuitive method which first conducts the uniformly random sampling of 2D roots on 2D matting masks of eyebrow images, then lifts 2D roots to 3D using the same method in~\cref{subsec:rootfinder}.
From~\cref{tab:compare_root}, it is clear to see a significant reduction of two metrics with different neighborhood radius $\varphi$. 
In our setting, the target face is normalized to $[-1, 1]$ in width. Thus, we choose $\varphi$ as 0.04, 0.02, and 0.01 for comprehensive evaluations.
Visual comparisons are illustrated by predicted 3D roots with the colorization of $DCD^{'}$ ($\varphi=0.02$) and extracted 2D roots with the background of ground-truth eyebrows (see~\cref{fig:root_visual}). The most distinguished problem of random sampling is that randomly sampled roots cannot cover the eyebrow shape with correct sparseness. Almost all selected roots fall in dense pixel regions and barely on sparse regions, such as the tails and the fibers between two eyebrows. In contrast, \rootFinder maintains the overall boundary very well, while also preserving the local distribution suggested by the input image.

\input{table/table_rootFinder_3D}

\input{figure/fig_root_vis}

\textbf{Evaluation of 3D orientation.} 
We follow \cite{wu2022neuralhdhair, zheng2023hairstep} to use $L_2$ error to evaluate the orientation field. The $L_2$ error of our method is 0.0951. From~\cref{fig:comparison}, the results of our method have a much better alignment with input eyebrow images in fiber growing direction than previous ruled-based approaches~\cite{herrera2010toward,rotger2019single}, indicating that the predicted orientation field is capable of describing how the eyebrow should grow in the aforementioned implicit manner. 

\textbf{Evaluation of fiber length.}
To evaluate the effectiveness of our method on the determination of fiber length, we design experiments with five settings: ending fiber growing with the mean length of all fibers in \dataset (baseline), cutting using the ground-truth contour mesh, our method without recurrent module, our method without positional encoding $\Phi(p)$, and our full model \fiberEnder. As for the mesh-cutting method, we follow~\cite {hu2017avatar} to first extract watertight ground-truth eyebrow meshes from ground-truth eyebrows using the method in~\cite{zhu2005animating}. 
Given the ground-truth mesh, we can easily stop the growth of a fiber if it grows out of the boundary of the mesh. 
The quantitative results are displayed in~\cref{tab:compare_length}. Our full model has a 45.77\% reduction in $MLE$ compared with the intuitive approach by setting all fibers into the mean length and 60.16\% less than the mesh-cutting method, even if it uses ground-truth meshes rather than predicted meshes.
Moreover, ablation of the recurrent architecture and positional information does affect the results to a certain extent, which can provide evidence to our assumption that historical growing features help in the learning of length control of eyebrow fibers. 
Visual results colorized by $MLE$ for each fiber in~\cref{fig:length_visual}, revealing more obvious differences between these approaches. Growing with mean length tends to preserve overall characteristics but lose length details in the region that is out of distribution. Mesh-cutting can only cut off fibers correctly which should stop near the boundary of the mesh but fails to work on fibers that should stop inside the mesh, especially in the middle region of the eyebrows. Moreover, results will be even worse if the predicted coarse meshes are used instead of the ground-truth ones. To sum up, \fiberEnder outperforms other methods to much extent.

\input{table/tab_fiberEnder_MLE}
\input{figure/fig_length_vis}

\subsection{Comparisons}
This subsection is to compare \ems with existing related approaches by quantitative and qualitative results.

\textbf{Comparison with rule-based facial hair reconstruction.}
Since there have been no existing learning-based attempts for 3D eyebrow modeling tasks, we implement rule-based methods in~\cite{herrera2010toward} and~\cite{rotger2019single} to make comparisons. On the testing set of \dataset, we report IoU and $FDO$ for reference in~\cref{tab:comparison}. With the help of \rootFinder, \oriPredictor and \fiberEnder, \ems outperforms the traditional methods in both metrics. We also collect some in-the-wild photos with an iPhone Pro Max 13 in our local area. As shown in~\cref{fig:comparison}, our method achieves state-of-the-art performance. Compared with previous methods, our reconstructed eyebrows have a more similar shape matching with original images and the fiber growing pattern is more natural, thanks to the intrinsic advantages of the data-driven approach by introducing 3D priors to single view reconstruction. Results in~\cref{fig:comparison} prove that our method trained on synthetic data also works for the input of real images with a non-negligible domain gap. However, we have to admit that, due to the lack of diversity in FaceScape~\cite{yang2020facescape} (most identities are Asian), our dataset \dataset may fail to cover all properties such as eyebrow types, skin tones, and fiber colors despite our efforts to increase the variety of \dataset.  Although this limitation may have a certain negative impact on the generalization capability of our method, the reconstructed results tested on the online public images~\cite{pexels} with different skin tones and fiber colors still look fine (\cref{fig:revision_diverse}).


It is worth mentioning that the accuracy of eyebrow reconstruction, especially 3D root prediction, is partially based on an off-the-shelf single-view face reconstruction method~\cite{yang2020facescape}. Even when only coarse face models are obtained sometimes in in-the-wild examples, the quality of eyebrow reconstruction is still visually satisfactory. The reason may be that our method mainly relies more on the projected growing hints.

\textbf{Comparison with multi-view facial hair reconstruction.}
To further illustrate the effectiveness of our single-view reconstruction framework, we compare \ems with a multi-view method~\cite{beeler2012coupled} by taking only a front-view image as input. From the result in~\cref{fig:comparison_multi-view}, although \ems may not produce highly detailed results like multi-view capture systems, it has already depicted the overall appearance of a comparable level with a more handy setting.  

\textbf{Comparison with single-view scalp hair reconstruction.}
To make the experiments more comprehensive, we compare our method with state-of-the-art single-view scalp hair modeling methods~\cite{wu2022neuralhdhair, zheng2023hairstep}, which are usually designed to grow fibers from dense pre-sampled roots on the scalp according to an implicit orientation field, and cut using predicted coarse meshes. For convenience, we compare \ems with cutting by ground-truth meshes, which is an even more tolerant setting. IoU and $DCD$ are reported in~\cref{tab:comparison}, where \ems is better thanks to module design for eyebrow characteristics especially.

\input{table/tab_comparison}
\input{figure/fig_comparison}
\input{figure/fig_revision_diverse}
\input{figure/fig_comparison_multi-view}

\subsection{Ablation Study}
To better demonstrate the necessity of the designing for each module, the whole \ems system is ablated into three settings: growing with random roots and a mean length (baseline), baseline with only \fiberEnder and baseline with only \rootFinder. They are all compared with each other and our full model by the metrics of IoU and $FDO$ against the ground truth on the testing set. 

\input{table/table_ablation}

From the comparisons of IoU in~\cref{tab:ablation} and~\cref{fig:ablation}, we can tell that \rootFinder exerts a significant effect on the recovery of the overall shape of 3D eyebrows. Using randomly sampled roots, the tails and the region between two eyebrows cannot be covered very well.
Furthermore, \fiberEnder also brings some bonus on both metrics, that is, controlling the length of every fiber not only preserves a better overall volume but also modifies wrong orientation by ending inappropriate growing. 
However, with the huge improvement of the IoU brought by \rootFinder, a small perturbation in the accuracy of orientation is acceptable.


\input{figure/fig_ablation}

\subsection{User Study}
Last but not least, we make a user study on three methods in~\cref{fig:comparison}, with a total of 56 users involved. In the questionnaire, they are provided with 8 randomly chosen reconstructed examples by different approaches and asked to select one result that matches the given close-up photos best from three randomly sorted options. 78.57\% of answers vote for \ems as the best, while 13.39\% and 8.04\%  vote for the methods of~\cite{herrera2010toward} and~\cite{rotger2019single}, respectively. It is supportive evidence for the superiority of our method.
\input{figure/fig_user_study}

%% file: table/table_rootFinder_3D.tex


\begin{table}[!h]
\begin{center}
\begin{tabular}{l|cc|cc|cc}
\hline
\multirow{2}{*}{Method} & $DCD$               & $NDE$             & $DCD$               & $NDE$             & $DCD$               & $NDE$            \\
                        & \multicolumn{2}{c|}{($\varphi$=0.04)} & \multicolumn{2}{c|}{($\varphi$=0.02)} & \multicolumn{2}{c}{($\varphi$=0.01)} \\ \hline

\emph{Random}                  & 0.2147            & 21.57           & 0.0811            & 7.503           & 0.03267           & 2.506          \\
\textbf{\rootFinder}              & 0.1261            & 15.13           & 0.04811           & 5.078           & 0.02222           & 1.735          \\ \hline
\end{tabular}
    \vspace{2mm}   
 \caption{Quantitative comparisons for the Localization of fiber roots.}
 \label{tab:compare_root}
\end{center}
\end{table}

%% file: figure/fig_root_vis.tex
\begin{figure*}[t]
\centering
\includegraphics[width=1.0\textwidth]{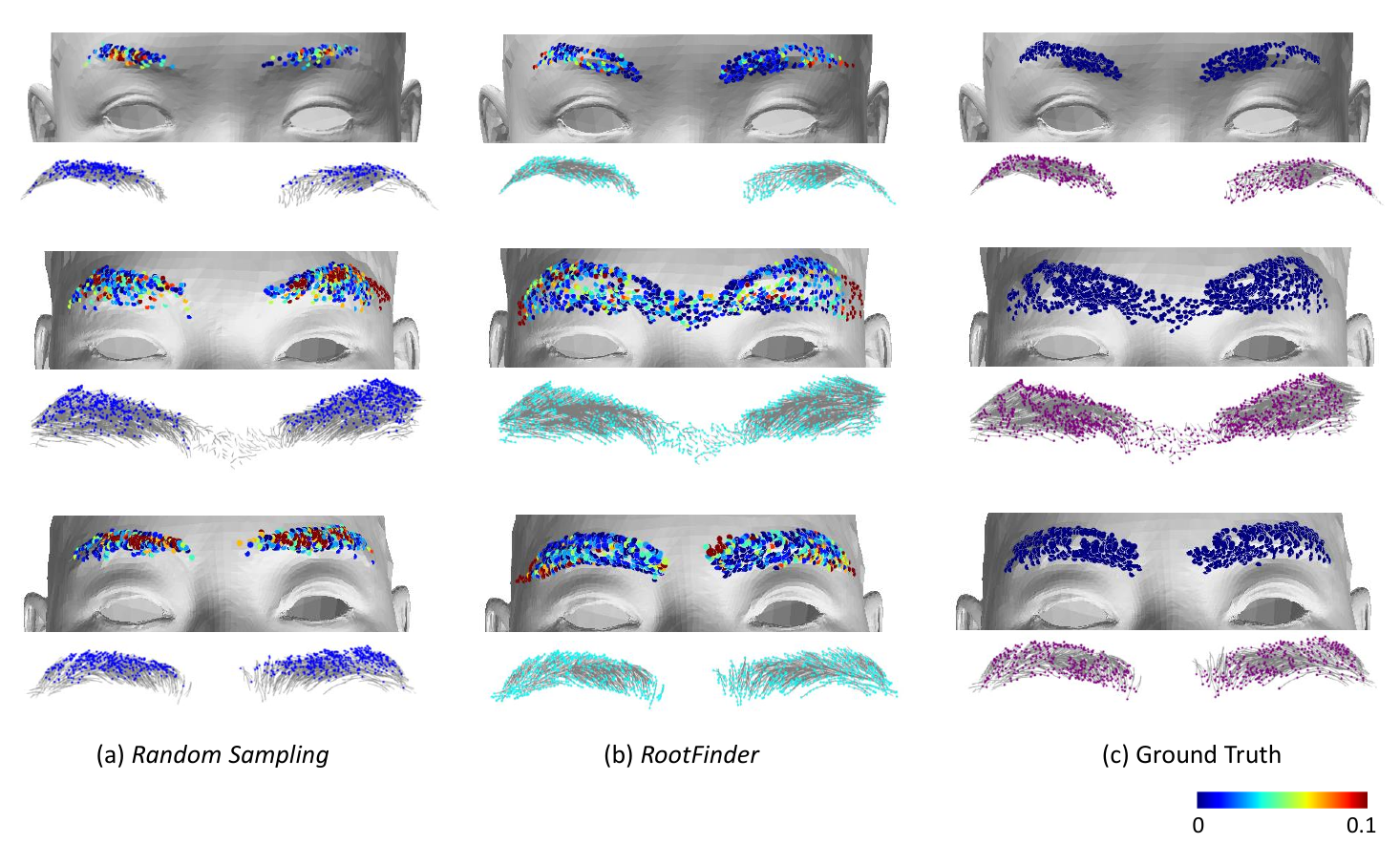}
\caption{Qualitative comparisons for the localization of fiber root. Every row presents the 3D root points and their corresponding projection in 2D of the same identity. From left to right:  
(a) Random sampling in eyebrow matting mask region,
(b) \rootFinder,
(c) Ground-truth. 
Each 3D root point is colorized according to $DCD^{'}$ ($\varphi=0.02$).
}
\label{fig:root_visual}
\end{figure*}

%% file: table/tab_fiberEnder_MLE.tex
\begin{table}[htbp]
 \begin{center}
  \begin{tabular}{l|c}
   \hline
   Method & $MLE$ ($\times10^{-2}$) $\downarrow$\\
        \hline
        \emph{End with mean length (Baseline)} & 2.9926 \\
        \emph{Cut by ground-truth mesh} & 4.0739\\
        {$\fiberEnder$} w/o recurrent module & 4.2130\\
        {$\fiberEnder$ w/o PE} & 1.8070 \textcolor{blue}{(-39.62\%)} \\
        \textbf{$\fiberEnder$} & 1.6229 \textcolor{blue}{(-45.77\%)} \\
  \hline
  \end{tabular}
  \vspace{3mm}   
  \caption{Quantitative comparisons for the determination of fiber length.}

 \label{tab:compare_length}
 \end{center}
\end{table}
\vspace{-3mm}   

%% file: figure/fig_length_vis.tex
\begin{figure*}[t]
\centering
\includegraphics[width=1.0\textwidth]{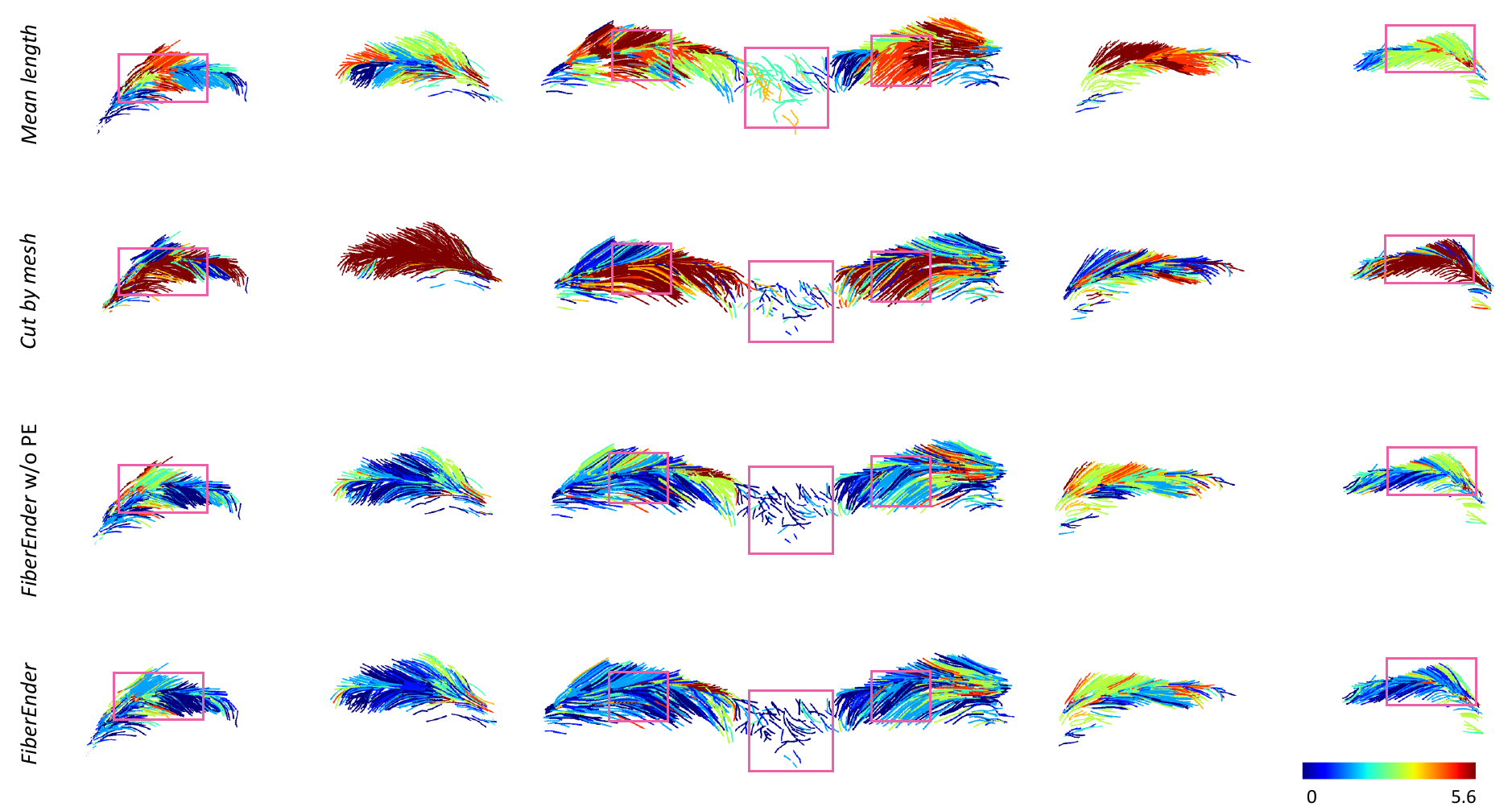}
 \vspace{1mm}   
\caption{Qualitative comparisons for the determination of fiber length.
From top to bottom, we show the reconstructed 3D eyebrow models of ending with mean length, cutting by ground-truth mesh, \fiberEnder w/o PE and \fiberEnder, respectively. Every column shows the results of the same identity.
Each fiber is colorized according to MLE ($\times10^{-2}$).}
\label{fig:length_visual}
\end{figure*}

%% file: table/tab_comparison.tex
\begin{table}[!h]
 \begin{center}
  \begin{tabular}{l|c c}
   \hline
   Method & IoU $\uparrow$ & $FDO$ $\downarrow$\\
        \hline
        {\cite{herrera2010toward}} & 0.5095 & 0.9752  \\
        {\cite{rotger2019single}} & 0.5126 & 0.9848  \\
        \emph{Scalp hair SVR} & 0.6160 & 0.2758 \\
         \textbf{\ems} & 0.8756 & 0.2480 \\

   \hline
  \end{tabular}
    \vspace{3mm}   
  \caption{Quantitative comparisons with existing related methods.}
 \label{tab:comparison}
 \end{center}
\end{table} 
\vspace{-3mm}

%% file: figure/fig_comparison.tex
\begin{figure*}[t]
\centering
\includegraphics[width=1.0\textwidth]{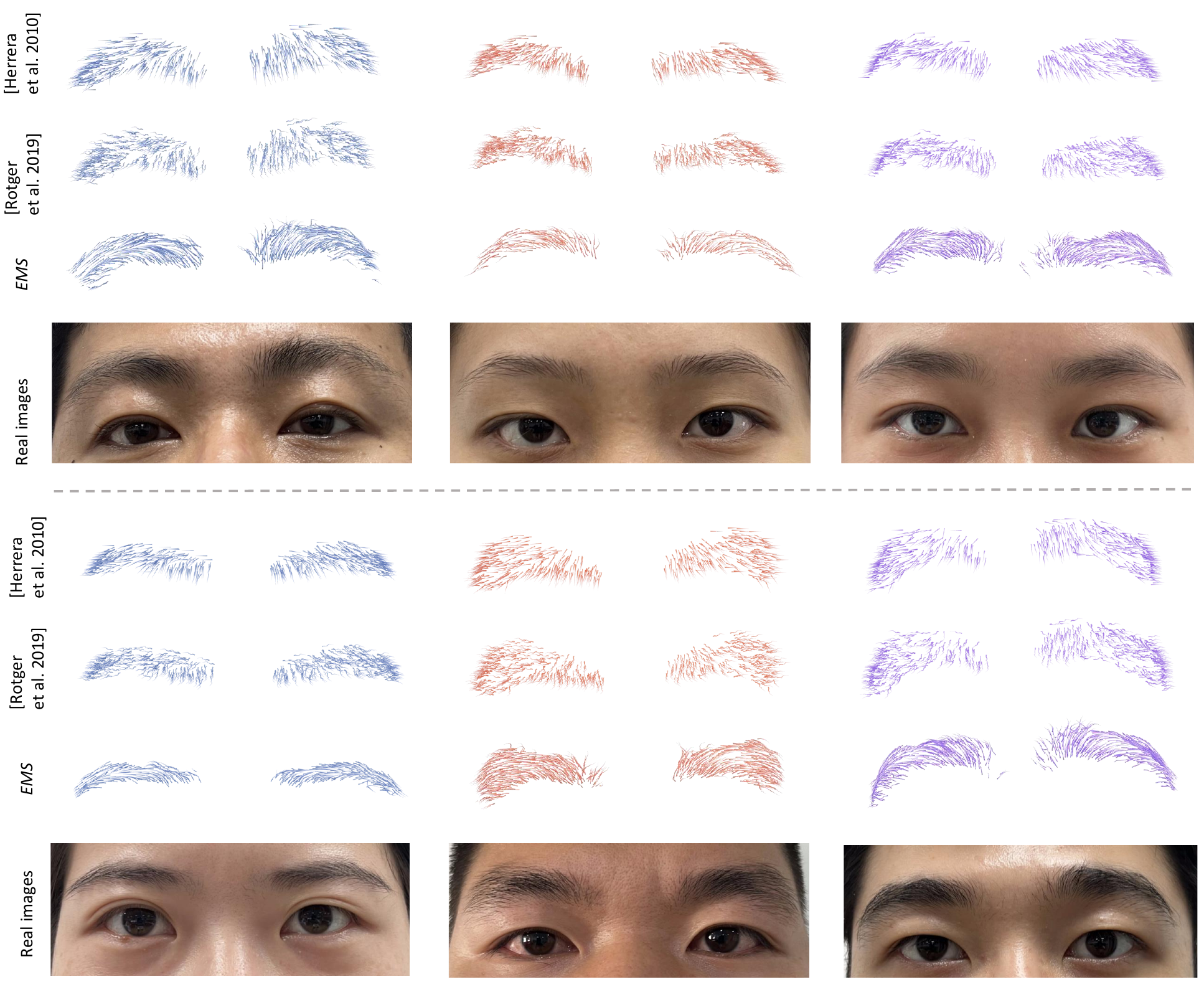}
\caption{Qualitative comparisons for in-the-wild real images.
From top to bottom, we show the reconstructed 3D eyebrow models of previous methods~\cite{herrera2010toward}, ~\cite{rotger2019single}, and \ems, respectively. The different colors used have no particular meaning, just for fruitful illustration of different examples.}
\label{fig:comparison}
\end{figure*}

%% file: figure/fig_revision_diverse.tex
\begin{figure*}[t]
\centering
\includegraphics[width=1.0\textwidth]{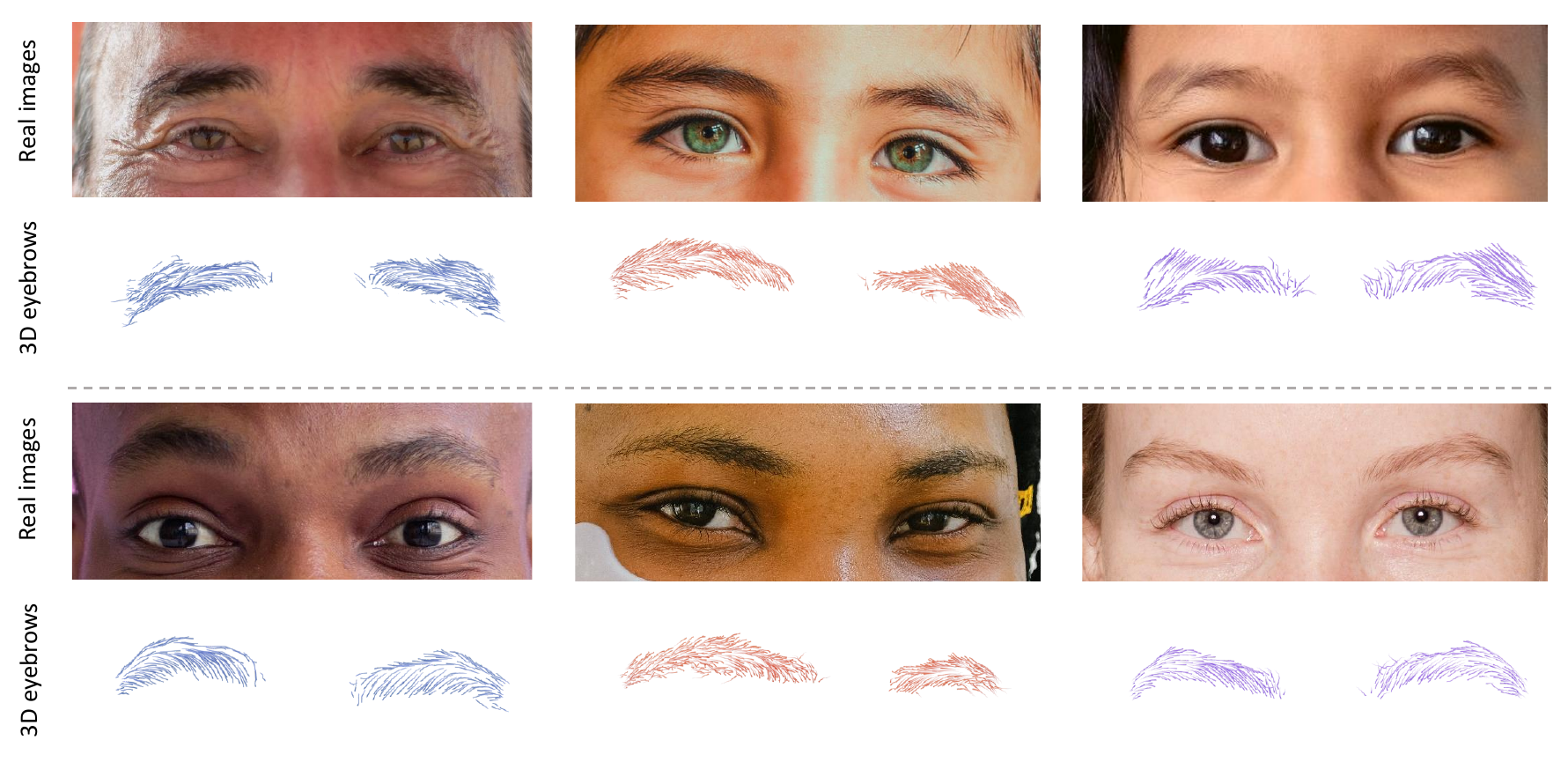}
\caption{Qualitative comparisons for diverse skin tones and fiber colors.
We show the reconstructed 3D eyebrow models and their corresponding original images. The different colors used have no particular meaning, just for fruitful illustration of different examples.}
\label{fig:revision_diverse}
\end{figure*}

%% file: figure/fig_comparison_multi-view.tex
\begin{figure}
\centering
\includegraphics[width=1.0\linewidth]{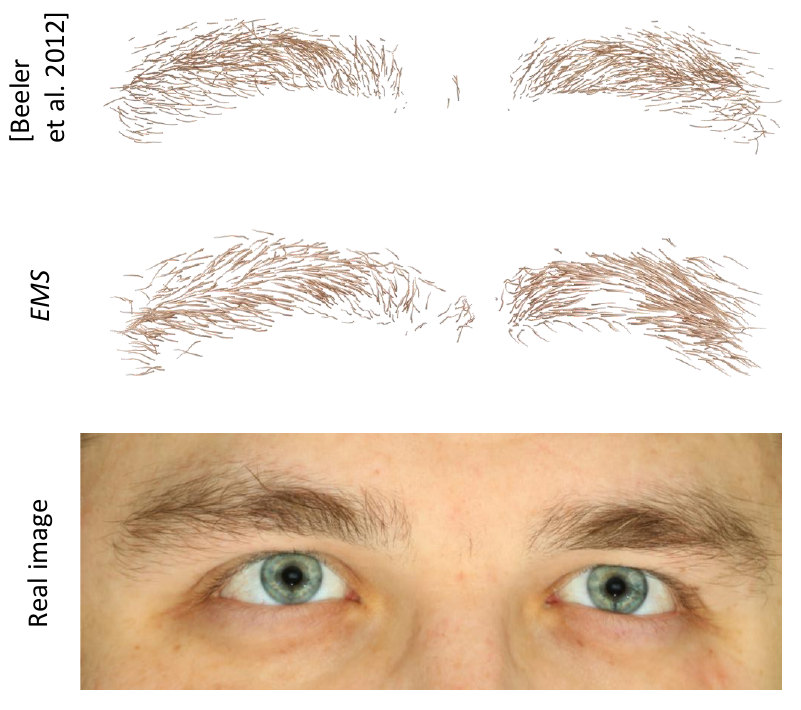}
\caption{Qualitative comparisons with the multi-view method. From top to bottom, we show the reconstructed 3D eyebrow models of \cite{beeler2012coupled}, \ems, and its corresponding real image.}
\label{fig:comparison_multi-view}
\end{figure}

%% file: table/table_ablation.tex
\begin{table}[!h]
 \begin{center}
  \begin{tabular}{l|c c}
   \hline
   Method & IoU $\uparrow$ & $FDO$ $\downarrow$\\
        \hline
        \emph{Baseline} & 0.5665& 0.2425  \\
        \emph{Baseline} $+$ \rootFinder & 0.8440& 0.2655  \\
        \emph{Baseline} $+$ \fiberEnder & 0.5918& 0.2338  \\
        \emph{\textbf{Full}} & 0.8756 & 0.2480 \\

   \hline
  \end{tabular}
    \vspace{3mm}   
  \caption{Quantitative comparisons for the ablation study.}
 \label{tab:ablation}
 \end{center}
\end{table} 

%% file: figure/fig_ablation.tex
\begin{figure*}[t]
\centering
\includegraphics[width=1.0\textwidth]{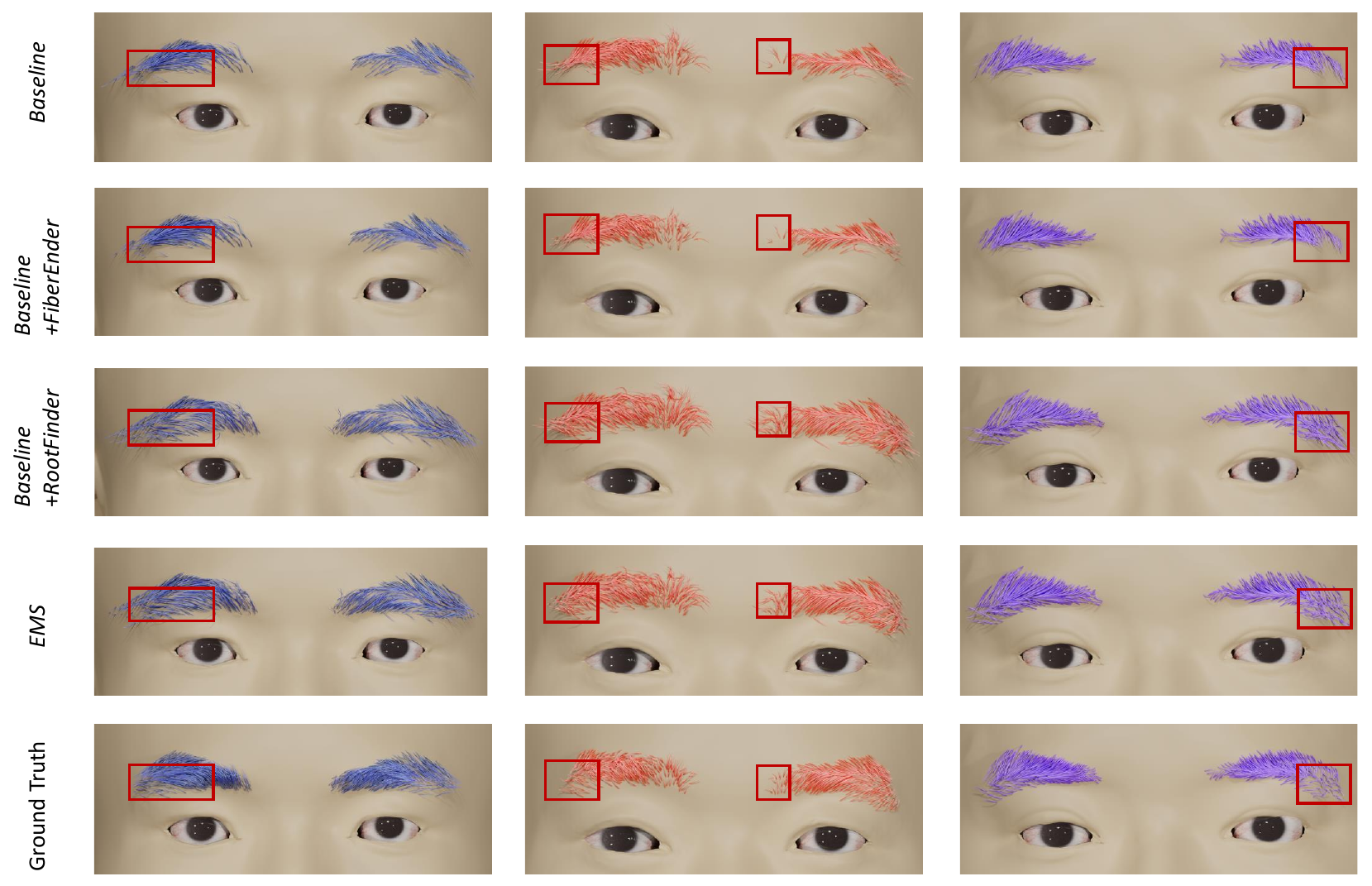}
\caption{Qualitative ablation study results. From top to bottom, we show the reconstructed 3D eyebrow models of the baseline, baseline with \fiberEnder only, baseline with \rootFinder only, our full model \ems and ground truth, respectively. The different colors used have no particular meaning, just for fruitful illustration of different examples.}
\label{fig:ablation}
\end{figure*}

%% file: figure/fig_user_study.tex
\begin{figure}
\centering
\includegraphics[width=0.7\linewidth]{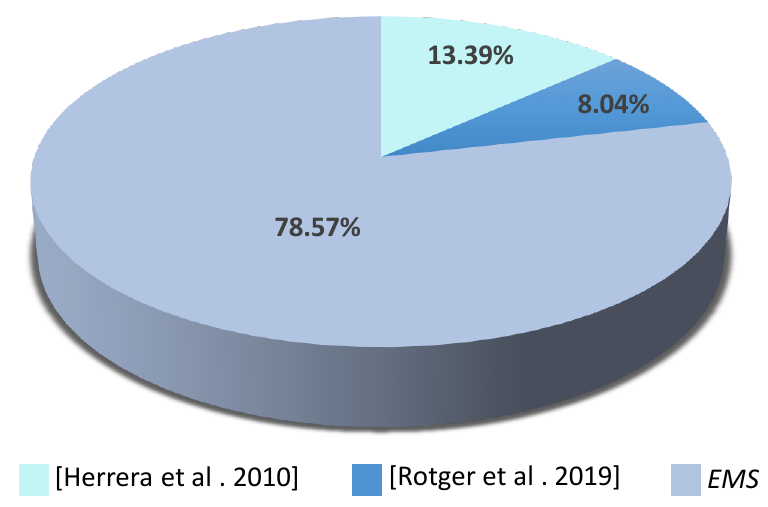}
\vspace{3mm}   
\caption{Statistics of user study. The pie chart displays the percentage of votes for the best eyebrow models among three methods represented by different colors.}
\label{fig:user_study}
\end{figure}

%% file: section/7_limitation_and_conclusion.tex
\section{Conclusion and Limitation }
To conclude, in this work, we propose the first learning-based system for fiber-level 3D eyebrow reconstruction from single-view images. \ems consists of three components \rootFinder, \oriPredictor and \fiberEnder, aiming to tackle the key challenges of eyebrow modeling. We also contribute the first high-quality 3D synthetic eyebrow dataset \dataset, which we use to facilitate the whole framework and introduce objective metrics to assess the performance of each module and overall reconstruction geometries. Experiments with visual and numerical results on both synthetic data and in-the-wild real images show the effectiveness of our system.

\textbf{Limitation.} 
Although \ems provides realistic reconstructed results on a wide range of data, it may fail in some cases shown in~\cref{fig:failure}: continuous dark eyebrow region of make-up or eyebrow tattoos (as the first row), occlusion of eyebrows by hair strands on the forehead (as the second row), and relatively slight contrast between skin tone and hair color (e.g., white eyebrows on light skin, as the third row). These artifacts are mainly caused by the eyebrow matting module~\cite{wang2022effective}. In these cases, the masks we obtain may be blurry or have wrong detection of eyebrow fibers, which will further lead to low-quality orientation maps, wrong distribution of eyebrow roots and inaccuracy of orientation and length prediction.
We believe that if we can find a more robust matting method or ease the dependence on it, our method is supposed to work better.
Besides, the generalization ability of our model is limited to eyebrows and may fail for other types of facial hairs, since prior knowledge learned from \dataset may be unsuitable to represent different growing rules. Another thing that needs to be noted is that our current model is trained using synthetic data only which takes the orientation map as the intermediate input. This makes the issue of domain gap still exist since the orientation maps extracted from real-world images are usually noisy. We leave this as a future research direction.    



\input{figure/fig_failure}

%% file: figure/fig_failure.tex
\begin{figure}
\centering
\includegraphics[width=1.0\linewidth]{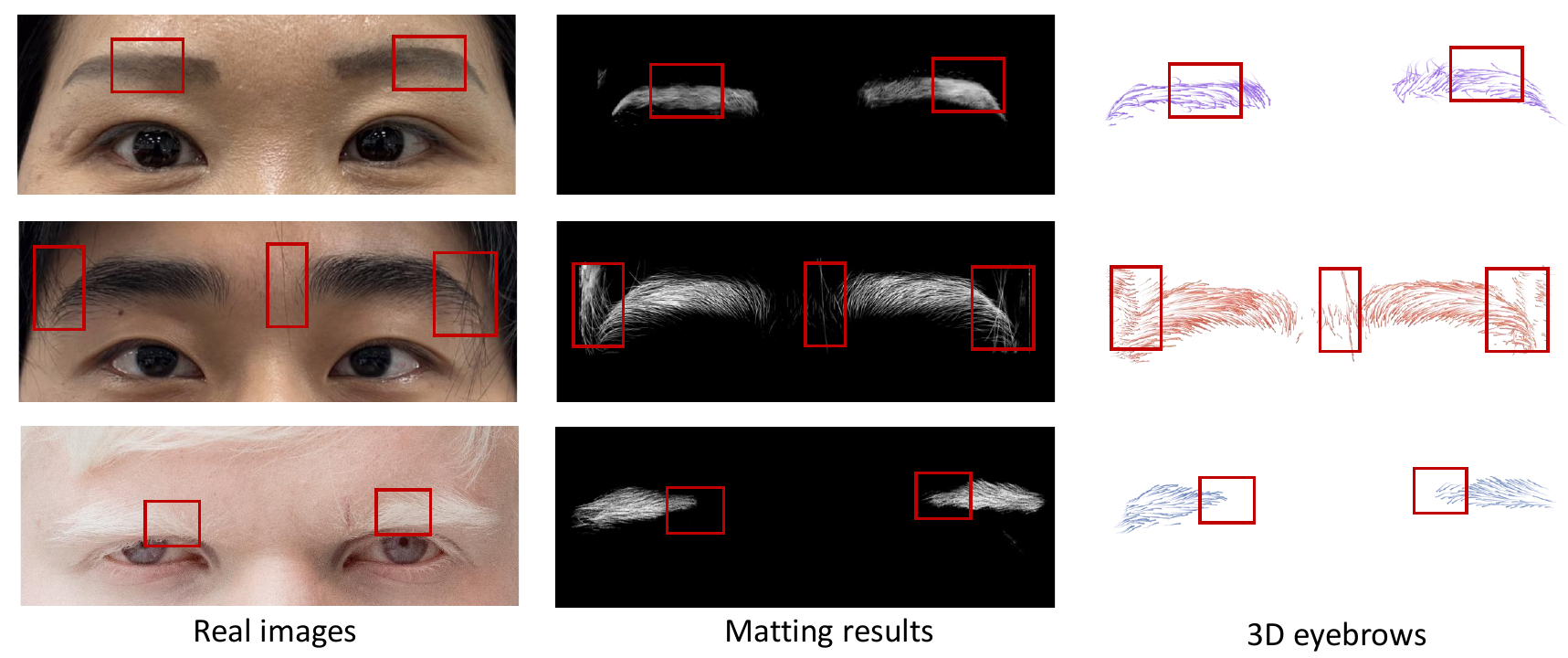}
\caption{Failure cases. From left to right, we show the real images, corresponding matting results and reconstructed 3D eyebrows.The different colors used have no particular meaning, just for fruitful illustration of different examples.}
\label{fig:failure}
\end{figure}

%% file: section/Acknowledgments.tex
\begin{acks}
The work was supported in part by NSFC with Grant No. 62293482, the Basic Research Project No. HZQB-KCZYZ-2021067 of Hetao ShenzhenHK S\&T Cooperation Zone, the National Key R\&D Program of China with Grant No. 2018YFB1800800, the Shenzhen Outstanding Talents Training Fund 202002, the Guangdong Research Projects No. 2017ZT07X152 and No. 2019CX01X104, the Guangdong Provincial Key Laboratory of Future Networks of Intelligence (Grant No. 2022B1212010001), and the Shenzhen Key Laboratory of Big Data and Artificial Intelligence (Grant No. ZDSYS201707251409055). It was also partially supported by the Hong Kong Research Grants Council under General Research Funds (HKU17206218). Additionally, it received partial support from NSFC-62172348, the Outstanding Young Fund of Guangdong Province with No. 2023B1515020055, and the Shenzhen General Project with No. JCYJ20220530143604010.
\end{acks}

%% file: Supp/section_Supp/1_implementation.tex
\section{Implementation Details}
In this section, we describe the details of our networks and training settings in \rootFinder, \oriPredictor and \fiberEnder experiments.

\textbf{\rootFinder}
We employ the U-Net~\cite{ronneberger2015u} to predict the density map of the input orientation image with the input size $1500 \times 600$.  Adam optimizer~\cite{kingma2014adam} is selected 
for training with the initial learning rate $1e^{-5}$ and decayed by the factor of 0.01. The network is trained with a batch size of 4 for 300 epochs on a single NVIDIA RTX-3090Ti GPU for roughly two days.

\textbf{\oriPredictor}
We use the Hourglass~\cite{newell2016stacked}  network with four stacks with the input size $1500\times600$ to extract local image features for learning the 3D orientation field. The layer sizes of MLP as the decoder are 277, 1024, 512, 256, 128, 3. Adam optimizer is used during training and the learning rate is set to $1e^{-4}$ and decayed by the factor of 0.1 in the $60^{th}$ epoch. The network is trained with a batch size of 4 for 100 epochs on a single NVIDIA RTX-3090Ti GPU for about 15 hours.

\textbf{\fiberEnder.}
\fiberEnder shares similar 2D  image features with \oriPredictor, so we load a pre-trained checkpoint to facilitate training. Despite using only one stack of feature lists for both the training and testing phases due to limited computing resources, we have achieved a high level of accuracy. All fibers of one eyebrow are passed into the network in a batch. As for the RNN encoder, the size of hidden code $h$ is $256\times1$. The layer sizes of MLP as the decoder are 277, 1024, 512, 256, 128, 1. Adam optimizer is used during training and the learning rate is initialized to $1e^{-4}$ at the beginning and decayed by the factor of 0.1 in the $50^{th}$ epoch. It takes about two days to train for 100 epochs on a single NVIDIA RTX-3090Ti GPU. It is worth mentioning that the growing directions inferred by \oriPredictor may not exactly align with the curves on the orientation maps during fiber synthesis. To narrow the domain gap between synthetic and real, we use fibers predicted by well-trained \oriPredictor growing from ground-truth eyebrow roots to train \fiberEnder. In other words, the labels used for loss calculation are actually pseudo-labels according to the length of the corresponding ground-truth fiber in the raw dataset.

%% file: Supp/section_Supp/2_dataset_length.tex
\section{Length Statistics of \dataset}

Our synthetic dataset \dataset enables the \ems system for training and evaluation. To better demonstrate data preparation for \fiberEnder experiments, we present length statistics for 400 models in \dataset as~\cref{tab:length_sta}. All fibers generated by artists consist of an equal number (20) of points for standardization. We take the longest fiber as a reference, calculate the growing step $\bar{s} = 0.014$ and regroup all fibers into different length levels according to $\bar{s}$, which are labeled as the ground truth during training. Length statistics for each level are presented in~\cref{tab:length_sta}. The mean length of fiber is 0.0714 (Level 5). Since the majority of fiber length concentrates in the middle classes, using the mean length to end fiber growing as our baseline for comparison is fair.

\input{Supp/table_Supp/tab_length_sta}

%% file: Supp/table_Supp/tab_length_sta.tex
\begin{table*}[!h]
 \begin{center}
  \begin{tabular}{l|c|c|c|c|c|c|c|c|c|c|c|c}
   \hline
    {Length level}&  1 & 2 & 3 & 4 & 5 & 6 & 7 & 8 & 9 & 10 & 11 & $\ge12$ \\
    \hline
    {Counts} & 881 & 4921 & 39884 & 143449 & 68229 & 50680 & 37236 & 22976 & 11159 & 5035 & 2341 & 2656\\
    {Percentages} & 0.23\% & 1.26\% & 10.24\% & 36.82\% & 17.51\% & 13.01\% & 9.56\% & 5.90\% & 2.86\% & 1.29\% & 0.60\% & 0.72\%\\

   \hline
  \end{tabular}
  \vspace{2mm}   
  \caption{Length statistics of \dataset with step $=0.014$.}
 \label{tab:length_sta}
 \end{center}
\end{table*} 

%% file: Supp/section_Supp/3_density_map_visualization.tex
\section{density map visualization}
\input{Supp/figure_Supp/fig_density_map}

We show the intermediate density map results of the \rootFinder module in \cref{fig:density_map}. The visual results reveal that the predicted density maps closely match the ground-truth density maps, enabling the accurate representation of diverse distributions of hair fiber roots.

%% file: Supp/figure_Supp/fig_density_map.tex
\begin{figure*}[t]
\centering
\includegraphics[width=0.9\textwidth]{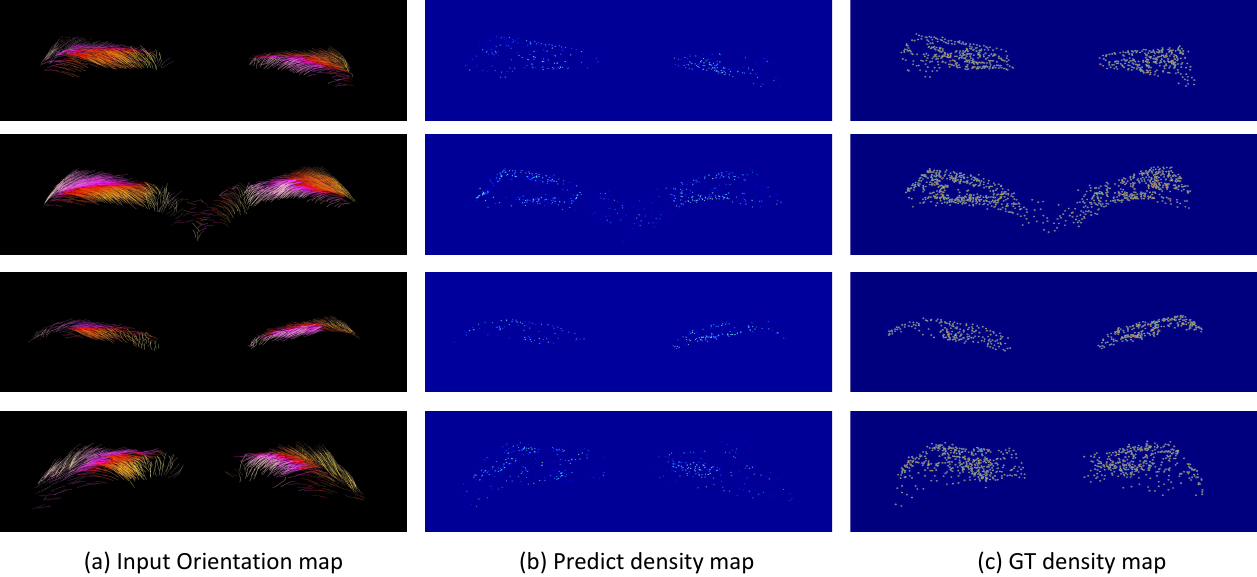}
  \vspace{-1mm}   
\caption{Qualitative results of the \rootFinder density map prediction. From left to right, we show the input orientation map, predict density map, and ground-truth density map, respectively.}
\label{fig:density_map}
\end{figure*}

%% file: Supp/section_Supp/4_more_visualizations_of_EBStore.tex
\section{More Visualizations of \dataset  Dataset}
We provide more visualizations of Facescape~\cite{yang2020facescape} front images and \dataset geometries for males, females and older people in~\cref{fig:more_dataset_male}, ~\cref{fig:more_dataset_female} and~\cref{fig:more_dataset_old}, respectively. These examples comprehensively show the diversity in eyebrow density, shape and growing pattern between different genders and ages.

\input{Supp/figure_Supp/fig_more_dataset_male}
\input{Supp/figure_Supp/fig_more_dataset_female}
\input{Supp/figure_Supp/fig_more_dataset_old}

%% file: Supp/figure_Supp/fig_more_dataset_male.tex
\begin{figure*}
\centering
\includegraphics[width=0.91\textwidth]{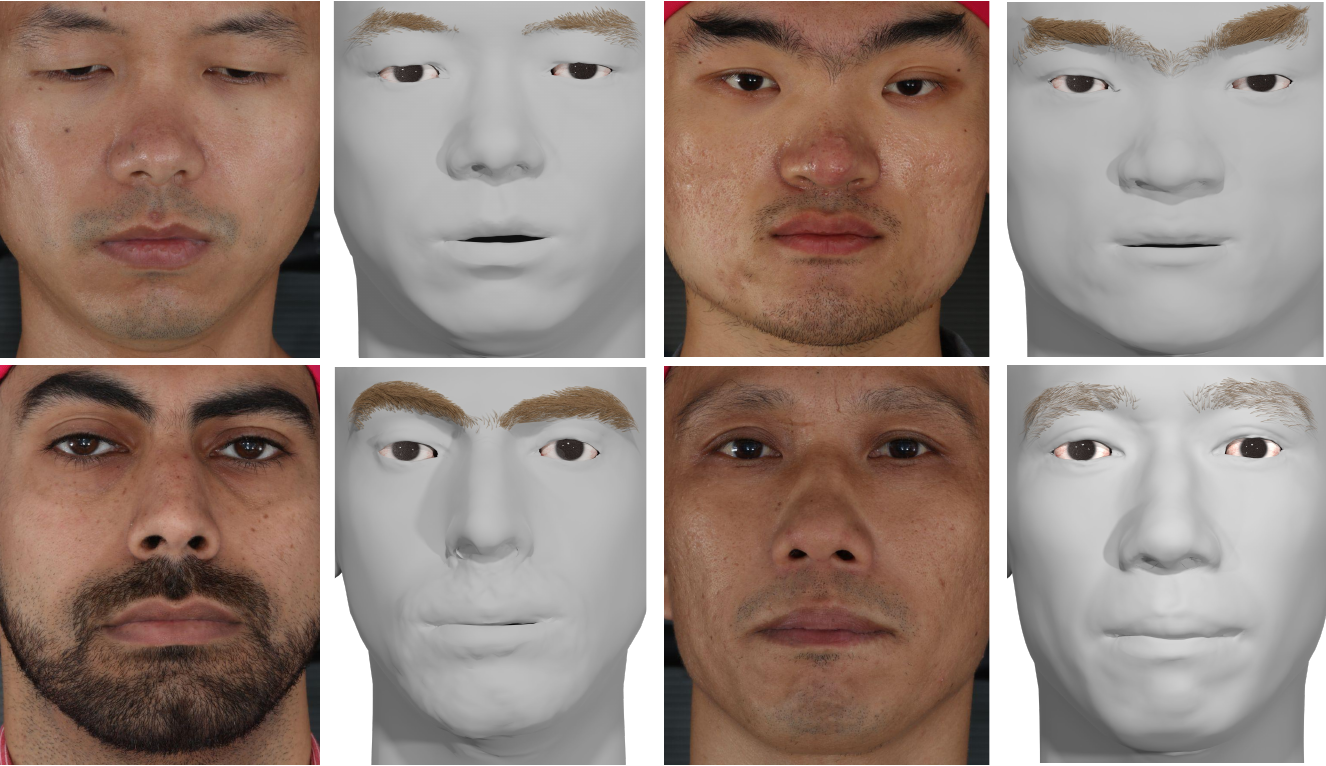}

\caption{\textbf{Male data samples of EBStore Dataset}.  More visualizations of Facescape front images (left)  and corresponding geometries (right) for males.}  
\label{fig:more_dataset_male}
\end{figure*}

%% file: Supp/figure_Supp/fig_more_dataset_female.tex
\begin{figure*}[t]
\centering
\includegraphics[width=0.9\textwidth]{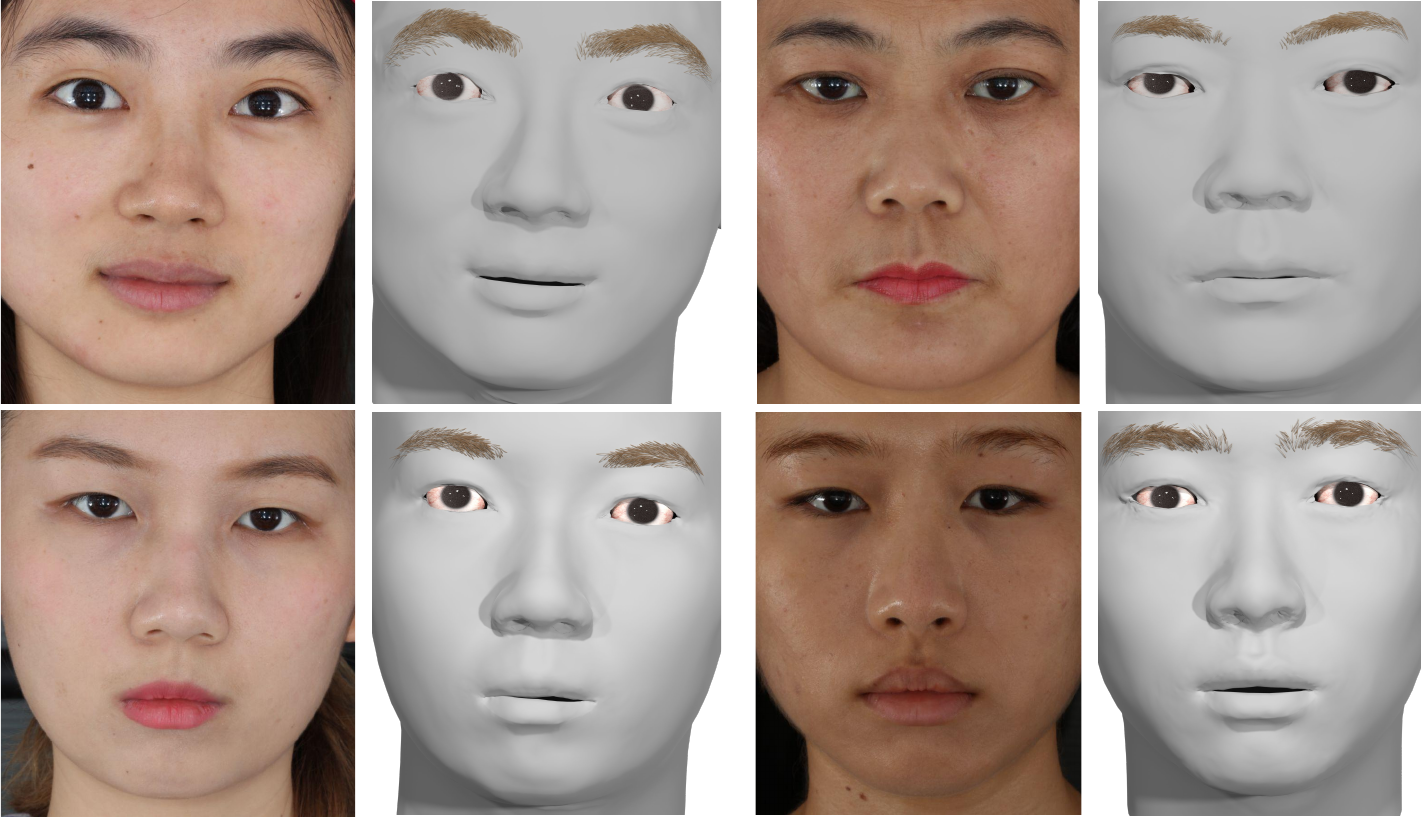}
  \vspace{-1mm}   
\caption{\textbf{Female data samples of EBStore Dataset}. More visualizations of Facescape front images (left)  and corresponding 3D geometries (right) for females without (the first row) and with (the second row) eyebrow make-up.} 
\label{fig:more_dataset_female}
\end{figure*}

%% file: Supp/figure_Supp/fig_more_dataset_old.tex
\begin{figure*}
\centering
\includegraphics[width=0.9\textwidth]{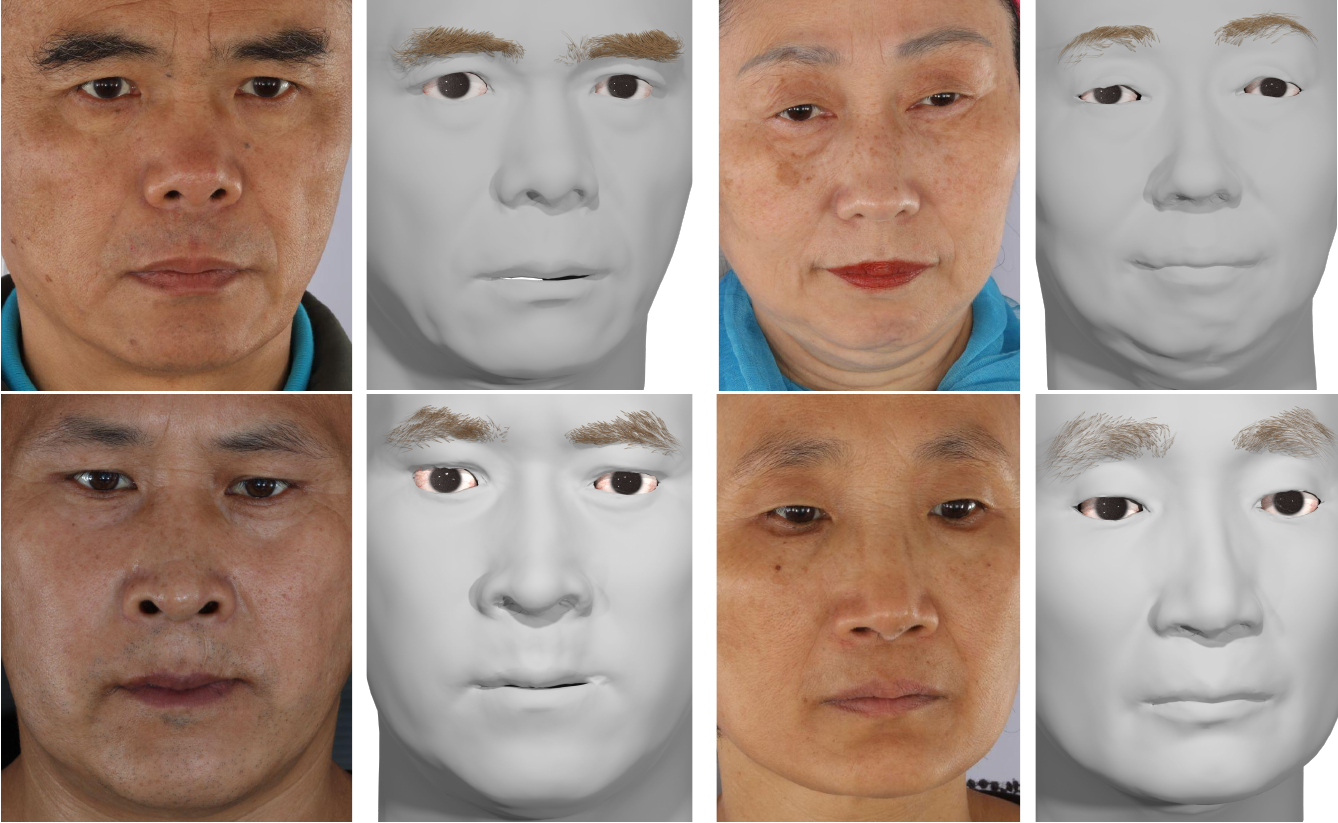}

  \vspace{-1mm}   
\caption{\textbf{Data samples of older individuals in the EBStore Dataset}. More visualizations of Facescape front images (left)  and corresponding 3D geometries (right) for old people. }  
\label{fig:more_dataset_old}
\end{figure*}

%% file: Supp/section_Supp/5_more_results_ablation.tex
\section{More Visual Results of Ablation Study}
We provide more visual results of the ablation study under four configurations: baseline of growing with random roots and a mean length, baseline with only \fiberEnder, baseline with only \rootFinder, and our \ems full model.  As illustrated in ~\cref{fig:ablation_sup}, \ems outperforms the ablated settings, with all modules working together. \rootFinder can capture the overall shape of an eyebrow, while \fiberEnder modifies the boundary details.
\input{Supp/figure_Supp/fig_ablation_sup}

%% file: Supp/figure_Supp/fig_ablation_sup.tex
\begin{figure*}[t]
\centering
\includegraphics[width=1.0\textwidth]{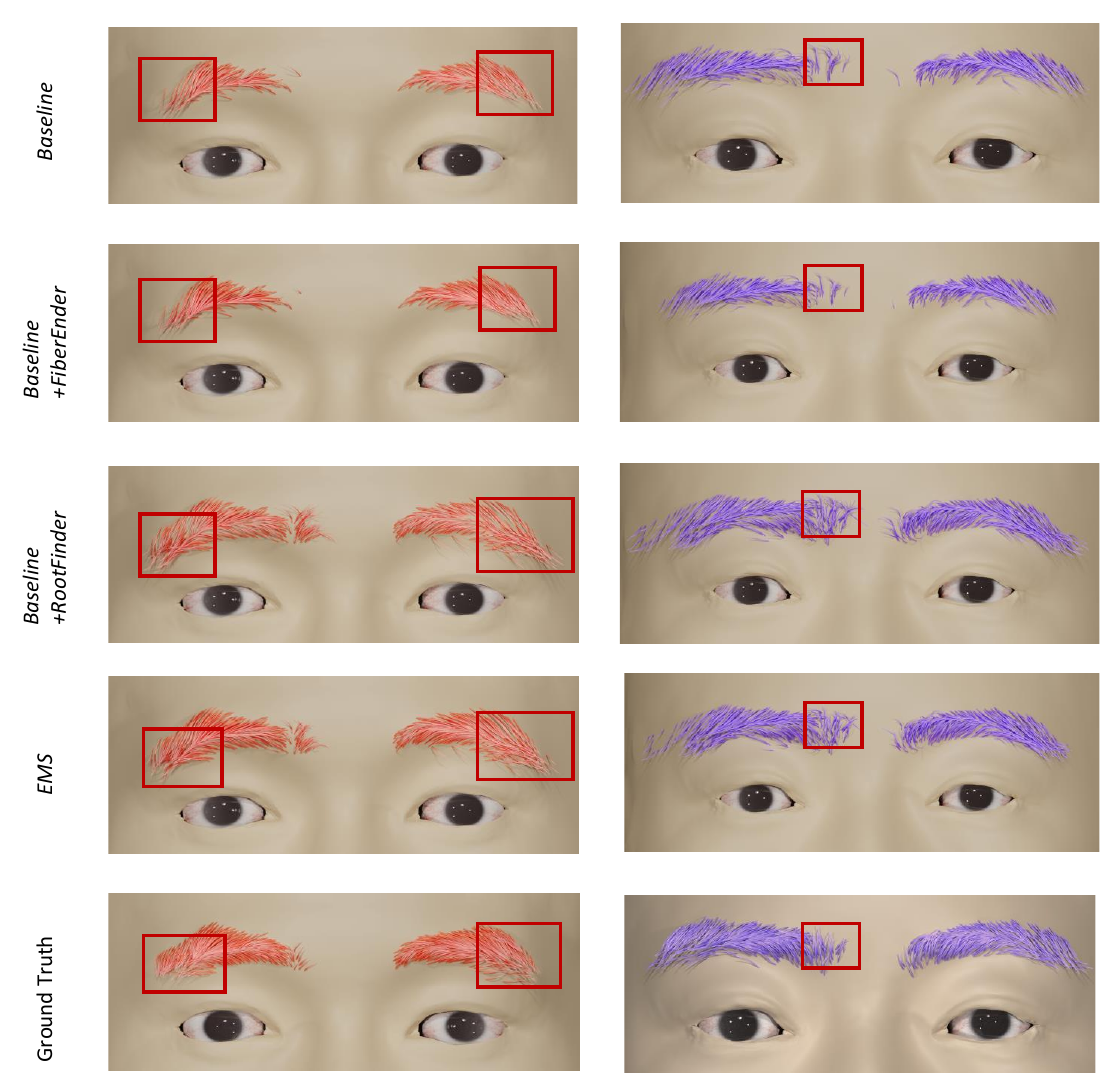}
\caption{Qualitative ablation study results. From top to bottom, we show the reconstructed 3D eyebrow models of the baseline, baseline with \fiberEnder only, baseline with \rootFinder only, our full model \ems, and ground truth, respectively. The different colors used have no particular meaning, just for fruitful illustration of different examples.}
\label{fig:ablation_sup}
\end{figure*}

%% file: acmtog-main.bbl

\begin{thebibliography}{83}


\ifx \showCODEN    \undefined \def \showCODEN     #1{\unskip}     \fi
\ifx \showDOI      \undefined \def \showDOI       #1{#1}\fi
\ifx \showISBNx    \undefined \def \showISBNx     #1{\unskip}     \fi
\ifx \showISBNxiii \undefined \def \showISBNxiii  #1{\unskip}     \fi
\ifx \showISSN     \undefined \def \showISSN      #1{\unskip}     \fi
\ifx \showLCCN     \undefined \def \showLCCN      #1{\unskip}     \fi
\ifx \shownote     \undefined \def \shownote      #1{#1}          \fi
\ifx \showarticletitle \undefined \def \showarticletitle #1{#1}   \fi
\ifx \showURL      \undefined \def \showURL       {\relax}        \fi
\providecommand\bibfield[2]{#2}
\providecommand\bibinfo[2]{#2}
\providecommand\natexlab[1]{#1}
\providecommand\showeprint[2][]{arXiv:#2}

\bibitem[Bao et~al\mbox{.}(2021)]%
        {bao2021high}
\bibfield{author}{\bibinfo{person}{Linchao Bao}, \bibinfo{person}{Xiangkai Lin}, \bibinfo{person}{Yajing Chen}, \bibinfo{person}{Haoxian Zhang}, \bibinfo{person}{Sheng Wang}, \bibinfo{person}{Xuefei Zhe}, \bibinfo{person}{Di Kang}, \bibinfo{person}{Haozhi Huang}, \bibinfo{person}{Xinwei Jiang}, \bibinfo{person}{Jue Wang}, {et~al\mbox{.}}} \bibinfo{year}{2021}\natexlab{}.
\newblock \showarticletitle{High-fidelity 3d digital human head creation from rgb-d selfies}.
\newblock \bibinfo{journal}{\emph{ACM Transactions on Graphics (TOG)}} \bibinfo{volume}{41}, \bibinfo{number}{1} (\bibinfo{year}{2021}), \bibinfo{pages}{1--21}.
\newblock


\bibitem[Beeler et~al\mbox{.}(2012)]%
        {beeler2012coupled}
\bibfield{author}{\bibinfo{person}{Thabo Beeler}, \bibinfo{person}{Bernd Bickel}, \bibinfo{person}{Gioacchino Noris}, \bibinfo{person}{Paul Beardsley}, \bibinfo{person}{Steve Marschner}, \bibinfo{person}{Robert~W Sumner}, {and} \bibinfo{person}{Markus Gross}.} \bibinfo{year}{2012}\natexlab{}.
\newblock \showarticletitle{Coupled 3D reconstruction of sparse facial hair and skin}.
\newblock \bibinfo{journal}{\emph{ACM Transactions on Graphics (TOG)}} \bibinfo{volume}{31}, \bibinfo{number}{4} (\bibinfo{year}{2012}).
\newblock


\bibitem[B{\'e}rard et~al\mbox{.}(2016)]%
        {berard2016lightweight}
\bibfield{author}{\bibinfo{person}{Pascal B{\'e}rard}, \bibinfo{person}{Derek Bradley}, \bibinfo{person}{Markus Gross}, {and} \bibinfo{person}{Thabo Beeler}.} \bibinfo{year}{2016}\natexlab{}.
\newblock \showarticletitle{Lightweight eye capture using a parametric model}.
\newblock \bibinfo{journal}{\emph{ACM Transactions on Graphics (TOG)}} \bibinfo{volume}{35}, \bibinfo{number}{4} (\bibinfo{year}{2016}), \bibinfo{pages}{1--12}.
\newblock


\bibitem[B{\'e}rard et~al\mbox{.}(2014)]%
        {berard2014high}
\bibfield{author}{\bibinfo{person}{Pascal B{\'e}rard}, \bibinfo{person}{Derek Bradley}, \bibinfo{person}{Maurizio Nitti}, \bibinfo{person}{Thabo Beeler}, {and} \bibinfo{person}{Markus~H Gross}.} \bibinfo{year}{2014}\natexlab{}.
\newblock \showarticletitle{High-quality capture of eyes.}
\newblock \bibinfo{journal}{\emph{ACM Trans. Graph.}} \bibinfo{volume}{33}, \bibinfo{number}{6} (\bibinfo{year}{2014}), \bibinfo{pages}{223--1}.
\newblock


\bibitem[Bermano et~al\mbox{.}(2015)]%
        {bermano2015detailed}
\bibfield{author}{\bibinfo{person}{Amit Bermano}, \bibinfo{person}{Thabo Beeler}, \bibinfo{person}{Yeara Kozlov}, \bibinfo{person}{Derek Bradley}, \bibinfo{person}{Bernd Bickel}, {and} \bibinfo{person}{Markus Gross}.} \bibinfo{year}{2015}\natexlab{}.
\newblock \showarticletitle{Detailed spatio-temporal reconstruction of eyelids}.
\newblock \bibinfo{journal}{\emph{ACM Transactions on Graphics (TOG)}} \bibinfo{volume}{34}, \bibinfo{number}{4} (\bibinfo{year}{2015}), \bibinfo{pages}{1--11}.
\newblock


\bibitem[Chai et~al\mbox{.}(2016)]%
        {chai2016autohair}
\bibfield{author}{\bibinfo{person}{Menglei Chai}, \bibinfo{person}{Tianjia Shao}, \bibinfo{person}{Hongzhi Wu}, \bibinfo{person}{Yanlin Weng}, {and} \bibinfo{person}{Kun Zhou}.} \bibinfo{year}{2016}\natexlab{}.
\newblock \showarticletitle{Autohair: Fully automatic hair modeling from a single image}.
\newblock \bibinfo{journal}{\emph{ACM Transactions on Graphics}} \bibinfo{volume}{35}, \bibinfo{number}{4} (\bibinfo{year}{2016}).
\newblock


\bibitem[Chai et~al\mbox{.}(2013)]%
        {chai2013dynamic}
\bibfield{author}{\bibinfo{person}{Menglei Chai}, \bibinfo{person}{Lvdi Wang}, \bibinfo{person}{Yanlin Weng}, \bibinfo{person}{Xiaogang Jin}, {and} \bibinfo{person}{Kun Zhou}.} \bibinfo{year}{2013}\natexlab{}.
\newblock \showarticletitle{Dynamic hair manipulation in images and videos}.
\newblock \bibinfo{journal}{\emph{ACM Transactions on Graphics (TOG)}} \bibinfo{volume}{32}, \bibinfo{number}{4} (\bibinfo{year}{2013}), \bibinfo{pages}{1--8}.
\newblock


\bibitem[Chai et~al\mbox{.}(2012)]%
        {chai2012single}
\bibfield{author}{\bibinfo{person}{Menglei Chai}, \bibinfo{person}{Lvdi Wang}, \bibinfo{person}{Yanlin Weng}, \bibinfo{person}{Yizhou Yu}, \bibinfo{person}{Baining Guo}, {and} \bibinfo{person}{Kun Zhou}.} \bibinfo{year}{2012}\natexlab{}.
\newblock \showarticletitle{Single-view hair modeling for portrait manipulation}.
\newblock \bibinfo{journal}{\emph{ACM Transactions on Graphics (TOG)}} \bibinfo{volume}{31}, \bibinfo{number}{4} (\bibinfo{year}{2012}).
\newblock


\bibitem[Cho et~al\mbox{.}(2014)]%
        {cho2014learning}
\bibfield{author}{\bibinfo{person}{Kyunghyun Cho}, \bibinfo{person}{Bart Van~Merri{\"e}nboer}, \bibinfo{person}{Caglar Gulcehre}, \bibinfo{person}{Dzmitry Bahdanau}, \bibinfo{person}{Fethi Bougares}, \bibinfo{person}{Holger Schwenk}, {and} \bibinfo{person}{Yoshua Bengio}.} \bibinfo{year}{2014}\natexlab{}.
\newblock \showarticletitle{Learning phrase representations using RNN encoder-decoder for statistical machine translation}.
\newblock \bibinfo{journal}{\emph{arXiv preprint arXiv:1406.1078}} (\bibinfo{year}{2014}).
\newblock


\bibitem[Community(2018)]%
        {blender}
\bibfield{author}{\bibinfo{person}{Blender~Online Community}.} \bibinfo{year}{2018}\natexlab{}.
\newblock \bibinfo{booktitle}{\emph{Blender - a 3D modelling and rendering package}}.
\newblock Blender Foundation, Stichting Blender Foundation, Amsterdam.
\newblock
\urldef\tempurl%
\url{http://www.blender.org}
\showURL{%
\tempurl}


\bibitem[Dataset(2022)]%
        {eyebrows_dataset}
\bibfield{author}{\bibinfo{person}{Dataset}.} \bibinfo{year}{2022}\natexlab{}.
\newblock \bibinfo{title}{Eyebrows Dataset}.
\newblock \bibinfo{howpublished}{\url{ https://universe.roboflow.com/dataset-0dpth/eyebrows }}.
\newblock
\urldef\tempurl%
\url{https://universe.roboflow.com/dataset-0dpth/eyebrows}
\showURL{%
\tempurl}
\newblock
\shownote{visited on 2023-05-12}.


\bibitem[Deng et~al\mbox{.}(2019)]%
        {deng2019accurate}
\bibfield{author}{\bibinfo{person}{Yu Deng}, \bibinfo{person}{Jiaolong Yang}, \bibinfo{person}{Sicheng Xu}, \bibinfo{person}{Dong Chen}, \bibinfo{person}{Yunde Jia}, {and} \bibinfo{person}{Xin Tong}.} \bibinfo{year}{2019}\natexlab{}.
\newblock \showarticletitle{Accurate 3d face reconstruction with weakly-supervised learning: From single image to image set}. In \bibinfo{booktitle}{\emph{Proceedings of the IEEE/CVF conference on computer vision and pattern recognition workshops}}. \bibinfo{pages}{0--0}.
\newblock


\bibitem[Dinev et~al\mbox{.}(2018)]%
        {dinev2018user}
\bibfield{author}{\bibinfo{person}{Dimitar Dinev}, \bibinfo{person}{Thabo Beeler}, \bibinfo{person}{Derek Bradley}, \bibinfo{person}{Moritz B{\"a}cher}, \bibinfo{person}{Hongyi Xu}, {and} \bibinfo{person}{Ladislav Kavan}.} \bibinfo{year}{2018}\natexlab{}.
\newblock \showarticletitle{User-guided lip correction for facial performance capture}. In \bibinfo{booktitle}{\emph{Computer Graphics Forum}}, Vol.~\bibinfo{volume}{37}. \bibinfo{pages}{93--101}.
\newblock


\bibitem[Ester et~al\mbox{.}(1996)]%
        {ester1996density}
\bibfield{author}{\bibinfo{person}{Martin Ester}, \bibinfo{person}{Hans-Peter Kriegel}, \bibinfo{person}{J{\"o}rg Sander}, \bibinfo{person}{Xiaowei Xu}, {et~al\mbox{.}}} \bibinfo{year}{1996}\natexlab{}.
\newblock \showarticletitle{A density-based algorithm for discovering clusters in large spatial databases with noise.}. In \bibinfo{booktitle}{\emph{kdd}}, Vol.~\bibinfo{volume}{96}. \bibinfo{pages}{226--231}.
\newblock


\bibitem[Feng et~al\mbox{.}(2021)]%
        {DECA:Siggraph2021}
\bibfield{author}{\bibinfo{person}{Yao Feng}, \bibinfo{person}{Haiwen Feng}, \bibinfo{person}{Michael~J. Black}, {and} \bibinfo{person}{Timo Bolkart}.} \bibinfo{year}{2021}\natexlab{}.
\newblock \showarticletitle{Learning an Animatable Detailed {3D} Face Model from In-The-Wild Images}.
\newblock \bibinfo{journal}{\emph{ACM Transactions on Graphics, (Proc. SIGGRAPH)}} \bibinfo{volume}{40}, \bibinfo{number}{8}.
\newblock
\urldef\tempurl%
\url{https://doi.org/10.1145/3450626.3459936}
\showURL{%
\tempurl}


\bibitem[Fyffe(2012)]%
        {fyffe2012high}
\bibfield{author}{\bibinfo{person}{Graham Fyffe}.} \bibinfo{year}{2012}\natexlab{}.
\newblock \showarticletitle{High fidelity facial hair capture}.
\newblock In \bibinfo{booktitle}{\emph{ACM SIGGRAPH 2012 Talks}}.
\newblock


\bibitem[Garrido et~al\mbox{.}(2016)]%
        {garrido2016corrective}
\bibfield{author}{\bibinfo{person}{Pablo Garrido}, \bibinfo{person}{Michael Zollh{\"o}fer}, \bibinfo{person}{Chenglei Wu}, \bibinfo{person}{Derek Bradley}, \bibinfo{person}{Patrick P{\'e}rez}, \bibinfo{person}{Thabo Beeler}, {and} \bibinfo{person}{Christian Theobalt}.} \bibinfo{year}{2016}\natexlab{}.
\newblock \showarticletitle{Corrective 3D reconstruction of lips from monocular video.}
\newblock \bibinfo{journal}{\emph{ACM Trans. Graph.}} \bibinfo{volume}{35}, \bibinfo{number}{6} (\bibinfo{year}{2016}), \bibinfo{pages}{219--1}.
\newblock


\bibitem[Giles et~al\mbox{.}(1994)]%
        {giles1994dynamic}
\bibfield{author}{\bibinfo{person}{C~Lee Giles}, \bibinfo{person}{Gary~M Kuhn}, {and} \bibinfo{person}{Ronald~J Williams}.} \bibinfo{year}{1994}\natexlab{}.
\newblock \showarticletitle{Dynamic recurrent neural networks: Theory and applications}.
\newblock \bibinfo{journal}{\emph{IEEE Transactions on Neural Networks}} \bibinfo{volume}{5}, \bibinfo{number}{2} (\bibinfo{year}{1994}), \bibinfo{pages}{153--156}.
\newblock


\bibitem[Grabli et~al\mbox{.}(2002)]%
        {grabli2002image}
\bibfield{author}{\bibinfo{person}{Stephane Grabli}, \bibinfo{person}{Francois~X. Sillion}, \bibinfo{person}{Stephen~R. Marschner}, {and} \bibinfo{person}{Jerome~E. Lengyel}.} \bibinfo{year}{2002}\natexlab{}.
\newblock \showarticletitle{Image-based hair capture by inverse lighting}. In \bibinfo{booktitle}{\emph{Proceedings of Graphics Interface (GI)}}. \bibinfo{pages}{51--58}.
\newblock


\bibitem[Guo et~al\mbox{.}(2020)]%
        {guo2020towards}
\bibfield{author}{\bibinfo{person}{Jianzhu Guo}, \bibinfo{person}{Xiangyu Zhu}, \bibinfo{person}{Yang Yang}, \bibinfo{person}{Fan Yang}, \bibinfo{person}{Zhen Lei}, {and} \bibinfo{person}{Stan~Z Li}.} \bibinfo{year}{2020}\natexlab{}.
\newblock \showarticletitle{Towards fast, accurate and stable 3d dense face alignment}. In \bibinfo{booktitle}{\emph{Computer Vision--ECCV 2020: 16th European Conference, Glasgow, UK, August 23--28, 2020, Proceedings, Part XIX}}. Springer, \bibinfo{pages}{152--168}.
\newblock


\bibitem[Hartigan et~al\mbox{.}(1979)]%
        {hartigan1979k}
\bibfield{author}{\bibinfo{person}{John~A Hartigan}, \bibinfo{person}{Manchek~A Wong}, {et~al\mbox{.}}} \bibinfo{year}{1979}\natexlab{}.
\newblock \showarticletitle{A k-means clustering algorithm}.
\newblock \bibinfo{journal}{\emph{Applied statistics}} \bibinfo{volume}{28}, \bibinfo{number}{1} (\bibinfo{year}{1979}), \bibinfo{pages}{100--108}.
\newblock


\bibitem[Herrera et~al\mbox{.}(2012)]%
        {herrera2012lighting}
\bibfield{author}{\bibinfo{person}{Tomas~Lay Herrera}, \bibinfo{person}{Arno Zinke}, {and} \bibinfo{person}{Andreas Weber}.} \bibinfo{year}{2012}\natexlab{}.
\newblock \showarticletitle{Lighting hair from the inside: A thermal approach to hair reconstruction}.
\newblock \bibinfo{journal}{\emph{ACM Transactions on Graphics (TOG)}} \bibinfo{volume}{31}, \bibinfo{number}{6} (\bibinfo{year}{2012}), \bibinfo{pages}{1--9}.
\newblock


\bibitem[Herrera et~al\mbox{.}(2010)]%
        {herrera2010toward}
\bibfield{author}{\bibinfo{person}{Tomas~Lay Herrera}, \bibinfo{person}{Arno Zinke}, \bibinfo{person}{Andreas Weber}, {and} \bibinfo{person}{Thomas Vetter}.} \bibinfo{year}{2010}\natexlab{}.
\newblock \showarticletitle{Toward image-based facial hair modeling}. In \bibinfo{booktitle}{\emph{Proceedings of the 26th Spring Conference on Computer Graphics}}.
\newblock


\bibitem[Hu et~al\mbox{.}(2014a)]%
        {hu2014robust}
\bibfield{author}{\bibinfo{person}{Liwen Hu}, \bibinfo{person}{Chongyang Ma}, \bibinfo{person}{Linjie Luo}, {and} \bibinfo{person}{Hao Li}.} \bibinfo{year}{2014}\natexlab{a}.
\newblock \showarticletitle{Robust hair capture using simulated examples}.
\newblock \bibinfo{journal}{\emph{ACM Transactions on Graphics (TOG)}} \bibinfo{volume}{33}, \bibinfo{number}{4} (\bibinfo{year}{2014}), \bibinfo{pages}{1--10}.
\newblock


\bibitem[Hu et~al\mbox{.}(2015)]%
        {hu2015single}
\bibfield{author}{\bibinfo{person}{Liwen Hu}, \bibinfo{person}{Chongyang Ma}, \bibinfo{person}{Linjie Luo}, {and} \bibinfo{person}{Hao Li}.} \bibinfo{year}{2015}\natexlab{}.
\newblock \showarticletitle{Single-view hair modeling using a hairstyle database}.
\newblock \bibinfo{journal}{\emph{ACM Transactions on Graphics (ToG)}} \bibinfo{volume}{34}, \bibinfo{number}{4} (\bibinfo{year}{2015}).
\newblock


\bibitem[Hu et~al\mbox{.}(2014b)]%
        {hu2014capturing}
\bibfield{author}{\bibinfo{person}{Liwen Hu}, \bibinfo{person}{Chongyang Ma}, \bibinfo{person}{Linjie Luo}, \bibinfo{person}{Li-Yi Wei}, {and} \bibinfo{person}{Hao Li}.} \bibinfo{year}{2014}\natexlab{b}.
\newblock \showarticletitle{Capturing braided hairstyles}.
\newblock \bibinfo{journal}{\emph{ACM Transactions on Graphics (TOG)}} \bibinfo{volume}{33}, \bibinfo{number}{6} (\bibinfo{year}{2014}).
\newblock


\bibitem[Hu et~al\mbox{.}(2017)]%
        {hu2017avatar}
\bibfield{author}{\bibinfo{person}{Liwen Hu}, \bibinfo{person}{Shunsuke Saito}, \bibinfo{person}{Lingyu Wei}, \bibinfo{person}{Koki Nagano}, \bibinfo{person}{Jaewoo Seo}, \bibinfo{person}{Jens Fursund}, \bibinfo{person}{Iman Sadeghi}, \bibinfo{person}{Carrie Sun}, \bibinfo{person}{Yen-Chun Chen}, {and} \bibinfo{person}{Hao Li}.} \bibinfo{year}{2017}\natexlab{}.
\newblock \showarticletitle{Avatar digitization from a single image for real-time rendering}.
\newblock \bibinfo{journal}{\emph{ACM Transactions on Graphics (ToG)}} \bibinfo{volume}{36}, \bibinfo{number}{6} (\bibinfo{year}{2017}), \bibinfo{pages}{1--14}.
\newblock


\bibitem[Jakob et~al\mbox{.}(2009)]%
        {jakob2009capturing}
\bibfield{author}{\bibinfo{person}{Wenzel Jakob}, \bibinfo{person}{Jonathan~T Moon}, {and} \bibinfo{person}{Steve Marschner}.} \bibinfo{year}{2009}\natexlab{}.
\newblock \showarticletitle{Capturing hair assemblies fiber by fiber}.
\newblock \bibinfo{journal}{\emph{ACM Transactions on Graphics(TOG)}} \bibinfo{volume}{28}, \bibinfo{number}{5} (\bibinfo{year}{2009}), \bibinfo{pages}{1--9}.
\newblock


\bibitem[Jiang et~al\mbox{.}(2019)]%
        {jiang2019crowd}
\bibfield{author}{\bibinfo{person}{Xiaolong Jiang}, \bibinfo{person}{Zehao Xiao}, \bibinfo{person}{Baochang Zhang}, \bibinfo{person}{Xiantong Zhen}, \bibinfo{person}{Xianbin Cao}, \bibinfo{person}{David Doermann}, {and} \bibinfo{person}{Ling Shao}.} \bibinfo{year}{2019}\natexlab{}.
\newblock \showarticletitle{Crowd counting and density estimation by trellis encoder-decoder networks}. In \bibinfo{booktitle}{\emph{Proceedings of the IEEE/CVF conference on computer vision and pattern recognition}}. \bibinfo{pages}{6133--6142}.
\newblock


\bibitem[Kingma and Ba(2014)]%
        {kingma2014adam}
\bibfield{author}{\bibinfo{person}{Diederik~P Kingma} {and} \bibinfo{person}{Jimmy Ba}.} \bibinfo{year}{2014}\natexlab{}.
\newblock \showarticletitle{Adam: A method for stochastic optimization}.
\newblock \bibinfo{journal}{\emph{arXiv preprint arXiv:1412.6980}} (\bibinfo{year}{2014}).
\newblock


\bibitem[Kuang et~al\mbox{.}(2022)]%
        {kuang2022deepmvshair}
\bibfield{author}{\bibinfo{person}{Zhiyi Kuang}, \bibinfo{person}{Yiyang Chen}, \bibinfo{person}{Hongbo Fu}, \bibinfo{person}{Kun Zhou}, {and} \bibinfo{person}{Youyi Zheng}.} \bibinfo{year}{2022}\natexlab{}.
\newblock \showarticletitle{DeepMVSHair: Deep Hair Modeling from Sparse Views}. In \bibinfo{booktitle}{\emph{ACM SIGGRAPH Asia 2022 Conference Papers}}.
\newblock


\bibitem[Lee et~al\mbox{.}(2020)]%
        {lee2020maskgan}
\bibfield{author}{\bibinfo{person}{Cheng-Han Lee}, \bibinfo{person}{Ziwei Liu}, \bibinfo{person}{Lingyun Wu}, {and} \bibinfo{person}{Ping Luo}.} \bibinfo{year}{2020}\natexlab{}.
\newblock \showarticletitle{Maskgan: Towards diverse and interactive facial image manipulation}. In \bibinfo{booktitle}{\emph{Proceedings of the IEEE/CVF Conference on Computer Vision and Pattern Recognition}}. \bibinfo{pages}{5549--5558}.
\newblock


\bibitem[LeGendre et~al\mbox{.}(2017)]%
        {legendre2017modeling}
\bibfield{author}{\bibinfo{person}{Chloe LeGendre}, \bibinfo{person}{Loc Hyunh}, \bibinfo{person}{Shanhe Wang}, {and} \bibinfo{person}{Paul Debevec}.} \bibinfo{year}{2017}\natexlab{}.
\newblock \showarticletitle{Modeling vellus facial hair from asperity scattering silhouettes}.
\newblock In \bibinfo{booktitle}{\emph{ACM SIGGRAPH 2017 Talks}}.
\newblock


\bibitem[Li et~al\mbox{.}(2022)]%
        {li2022eyenerf}
\bibfield{author}{\bibinfo{person}{Gengyan Li}, \bibinfo{person}{Abhimitra Meka}, \bibinfo{person}{Franziska Mueller}, \bibinfo{person}{Marcel~C Buehler}, \bibinfo{person}{Otmar Hilliges}, {and} \bibinfo{person}{Thabo Beeler}.} \bibinfo{year}{2022}\natexlab{}.
\newblock \showarticletitle{EyeNeRF: a hybrid representation for photorealistic synthesis, animation and relighting of human eyes}.
\newblock \bibinfo{journal}{\emph{ACM Transactions on Graphics (TOG)}} \bibinfo{volume}{41}, \bibinfo{number}{4} (\bibinfo{year}{2022}), \bibinfo{pages}{1--16}.
\newblock


\bibitem[Li et~al\mbox{.}(2017)]%
        {FLAME:SiggraphAsia2017}
\bibfield{author}{\bibinfo{person}{Tianye Li}, \bibinfo{person}{Timo Bolkart}, \bibinfo{person}{Michael.~J. Black}, \bibinfo{person}{Hao Li}, {and} \bibinfo{person}{Javier Romero}.} \bibinfo{year}{2017}\natexlab{}.
\newblock \showarticletitle{Learning a model of facial shape and expression from {4D} scans}.
\newblock \bibinfo{journal}{\emph{ACM Transactions on Graphics, (Proc. SIGGRAPH Asia)}} \bibinfo{volume}{36}, \bibinfo{number}{6} (\bibinfo{year}{2017}), \bibinfo{pages}{194:1--194:17}.
\newblock
\urldef\tempurl%
\url{https://doi.org/10.1145/3130800.3130813}
\showURL{%
\tempurl}


\bibitem[Liang et~al\mbox{.}(2018)]%
        {liang2018video}
\bibfield{author}{\bibinfo{person}{Shu Liang}, \bibinfo{person}{Xiufeng Huang}, \bibinfo{person}{Xianyu Meng}, \bibinfo{person}{Kunyao Chen}, \bibinfo{person}{Linda~G Shapiro}, {and} \bibinfo{person}{Ira Kemelmacher-Shlizerman}.} \bibinfo{year}{2018}\natexlab{}.
\newblock \showarticletitle{Video to fully automatic 3d hair model}.
\newblock \bibinfo{journal}{\emph{ACM Transactions on Graphics (TOG)}} \bibinfo{volume}{37}, \bibinfo{number}{6} (\bibinfo{year}{2018}).
\newblock


\bibitem[Liu et~al\mbox{.}(2019)]%
        {liu2019context}
\bibfield{author}{\bibinfo{person}{Weizhe Liu}, \bibinfo{person}{Mathieu Salzmann}, {and} \bibinfo{person}{Pascal Fua}.} \bibinfo{year}{2019}\natexlab{}.
\newblock \showarticletitle{Context-aware crowd counting}. In \bibinfo{booktitle}{\emph{Proceedings of the IEEE/CVF conference on computer vision and pattern recognition}}. \bibinfo{pages}{5099--5108}.
\newblock


\bibitem[Loper et~al\mbox{.}(2015)]%
        {SMPL:2015}
\bibfield{author}{\bibinfo{person}{Matthew Loper}, \bibinfo{person}{Naureen Mahmood}, \bibinfo{person}{Javier Romero}, \bibinfo{person}{Gerard Pons-Moll}, {and} \bibinfo{person}{Michael~J. Black}.} \bibinfo{year}{2015}\natexlab{}.
\newblock \showarticletitle{{SMPL}: A Skinned Multi-Person Linear Model}.
\newblock \bibinfo{journal}{\emph{ACM Trans. Graphics (Proc. SIGGRAPH Asia)}} \bibinfo{volume}{34}, \bibinfo{number}{6} (\bibinfo{date}{Oct.} \bibinfo{year}{2015}), \bibinfo{pages}{248:1--248:16}.
\newblock


\bibitem[Luo et~al\mbox{.}(2012)]%
        {luo2012multi}
\bibfield{author}{\bibinfo{person}{Linjie Luo}, \bibinfo{person}{Hao Li}, \bibinfo{person}{Sylvain Paris}, \bibinfo{person}{Thibaut Weise}, \bibinfo{person}{Mark Pauly}, {and} \bibinfo{person}{Szymon Rusinkiewicz}.} \bibinfo{year}{2012}\natexlab{}.
\newblock \showarticletitle{Multi-view hair capture using orientation fields}. In \bibinfo{booktitle}{\emph{2012 IEEE Conference on Computer Vision and Pattern Recognition}}. IEEE, \bibinfo{pages}{1490--1497}.
\newblock


\bibitem[Luo et~al\mbox{.}(2013)]%
        {luo2013structure}
\bibfield{author}{\bibinfo{person}{Linjie Luo}, \bibinfo{person}{Hao Li}, {and} \bibinfo{person}{Szymon Rusinkiewicz}.} \bibinfo{year}{2013}\natexlab{}.
\newblock \showarticletitle{Structure-aware hair capture}.
\newblock \bibinfo{journal}{\emph{ACM Transactions on Graphics (TOG)}} \bibinfo{volume}{32}, \bibinfo{number}{4} (\bibinfo{year}{2013}).
\newblock


\bibitem[Mildenhall et~al\mbox{.}(2021)]%
        {mildenhall2021nerf}
\bibfield{author}{\bibinfo{person}{Ben Mildenhall}, \bibinfo{person}{Pratul~P Srinivasan}, \bibinfo{person}{Matthew Tancik}, \bibinfo{person}{Jonathan~T Barron}, \bibinfo{person}{Ravi Ramamoorthi}, {and} \bibinfo{person}{Ren Ng}.} \bibinfo{year}{2021}\natexlab{}.
\newblock \showarticletitle{Nerf: Representing scenes as neural radiance fields for view synthesis}.
\newblock \bibinfo{journal}{\emph{Commun. ACM}} \bibinfo{volume}{65}, \bibinfo{number}{1} (\bibinfo{year}{2021}), \bibinfo{pages}{99--106}.
\newblock


\bibitem[Nam et~al\mbox{.}(2019)]%
        {nam2019strand}
\bibfield{author}{\bibinfo{person}{Giljoo Nam}, \bibinfo{person}{Chenglei Wu}, \bibinfo{person}{Min~H Kim}, {and} \bibinfo{person}{Yaser Sheikh}.} \bibinfo{year}{2019}\natexlab{}.
\newblock \showarticletitle{Strand-accurate multi-view hair capture}. In \bibinfo{booktitle}{\emph{Proceedings of the IEEE/CVF Conference on Computer Vision and Pattern Recognition}}. \bibinfo{pages}{155--164}.
\newblock


\bibitem[Newell et~al\mbox{.}(2016)]%
        {newell2016stacked}
\bibfield{author}{\bibinfo{person}{Alejandro Newell}, \bibinfo{person}{Kaiyu Yang}, {and} \bibinfo{person}{Jia Deng}.} \bibinfo{year}{2016}\natexlab{}.
\newblock \showarticletitle{Stacked hourglass networks for human pose estimation}. In \bibinfo{booktitle}{\emph{Computer Vision--ECCV 2016: 14th European Conference, Amsterdam, The Netherlands, October 11-14, 2016, Proceedings, Part VIII 14}}. Springer, \bibinfo{pages}{483--499}.
\newblock


\bibitem[Paris et~al\mbox{.}(2004)]%
        {paris2004capture}
\bibfield{author}{\bibinfo{person}{Sylvain Paris}, \bibinfo{person}{Hector~M. Briceno}, {and} \bibinfo{person}{Francois~X. Sillion}.} \bibinfo{year}{2004}\natexlab{}.
\newblock \showarticletitle{Capture of hair geometry from multiple images}.
\newblock \bibinfo{journal}{\emph{ACM transactions on graphics (TOG)}} \bibinfo{volume}{23}, \bibinfo{number}{3} (\bibinfo{year}{2004}), \bibinfo{pages}{712--719}.
\newblock


\bibitem[Paris et~al\mbox{.}(2008)]%
        {paris2008hair}
\bibfield{author}{\bibinfo{person}{Sylvain Paris}, \bibinfo{person}{Will Chang}, \bibinfo{person}{Oleg~I Kozhushnyan}, \bibinfo{person}{Wojciech Jarosz}, \bibinfo{person}{Wojciech Matusik}, \bibinfo{person}{Matthias Zwicker}, {and} \bibinfo{person}{Fr{\'e}do Durand}.} \bibinfo{year}{2008}\natexlab{}.
\newblock \showarticletitle{Hair photobooth: geometric and photometric acquisition of real hairstyles.}
\newblock \bibinfo{journal}{\emph{ACM Trans. Graph(Proc. SIGGRAPH)}} \bibinfo{volume}{27}, \bibinfo{number}{3} (\bibinfo{year}{2008}), \bibinfo{pages}{30}.
\newblock


\bibitem[Pexels(2021)]%
        {pexels}
\bibfield{author}{\bibinfo{person}{Pexels}.} \bibinfo{year}{2021}\natexlab{}.
\newblock \bibinfo{booktitle}{\emph{Pexels - The best free stock photos, royalty free images \& videos shared by creators}}.
\newblock
\urldef\tempurl%
\url{https://www.pexels.com}
\showURL{%
\tempurl}


\bibitem[Ploumpis et~al\mbox{.}(2022)]%
        {ploumpis20223d}
\bibfield{author}{\bibinfo{person}{Stylianos Ploumpis}, \bibinfo{person}{Stylianos Moschoglou}, \bibinfo{person}{Vasileios Triantafyllou}, {and} \bibinfo{person}{Stefanos Zafeiriou}.} \bibinfo{year}{2022}\natexlab{}.
\newblock \showarticletitle{3D human tongue reconstruction from single" in-the-wild" images}. In \bibinfo{booktitle}{\emph{Proceedings of the IEEE/CVF Conference on Computer Vision and Pattern Recognition}}. \bibinfo{pages}{2771--2780}.
\newblock


\bibitem[Ranjan et~al\mbox{.}(2021)]%
        {ranjan2021learning}
\bibfield{author}{\bibinfo{person}{Viresh Ranjan}, \bibinfo{person}{Udbhav Sharma}, \bibinfo{person}{Thu Nguyen}, {and} \bibinfo{person}{Minh Hoai}.} \bibinfo{year}{2021}\natexlab{}.
\newblock \showarticletitle{Learning to count everything}. In \bibinfo{booktitle}{\emph{Proceedings of the IEEE/CVF Conference on Computer Vision and Pattern Recognition}}. \bibinfo{pages}{3394--3403}.
\newblock


\bibitem[Ronneberger et~al\mbox{.}(2015)]%
        {ronneberger2015u}
\bibfield{author}{\bibinfo{person}{Olaf Ronneberger}, \bibinfo{person}{Philipp Fischer}, {and} \bibinfo{person}{Thomas Brox}.} \bibinfo{year}{2015}\natexlab{}.
\newblock \showarticletitle{U-net: Convolutional networks for biomedical image segmentation}. In \bibinfo{booktitle}{\emph{Medical Image Computing and Computer-Assisted Intervention--MICCAI 2015: 18th International Conference, Munich, Germany, October 5-9, 2015, Proceedings, Part III 18}}. Springer, \bibinfo{pages}{234--241}.
\newblock


\bibitem[Rosu et~al\mbox{.}(2022)]%
        {rosu2022neural}
\bibfield{author}{\bibinfo{person}{Radu~Alexandru Rosu}, \bibinfo{person}{Shunsuke Saito}, \bibinfo{person}{Ziyan Wang}, \bibinfo{person}{Chenglei Wu}, \bibinfo{person}{Sven Behnke}, {and} \bibinfo{person}{Giljoo Nam}.} \bibinfo{year}{2022}\natexlab{}.
\newblock \showarticletitle{Neural strands: Learning hair geometry and appearance from multi-view images}. In \bibinfo{booktitle}{\emph{European Conference on Computer Vision}}. Springer, \bibinfo{pages}{73--89}.
\newblock


\bibitem[Rotger et~al\mbox{.}(2019)]%
        {rotger2019single}
\bibfield{author}{\bibinfo{person}{Gemma Rotger}, \bibinfo{person}{Francesc Moreno-Noguer}, \bibinfo{person}{Felipe Lumbreras}, {and} \bibinfo{person}{Antonio Agudo}.} \bibinfo{year}{2019}\natexlab{}.
\newblock \showarticletitle{Single view facial hair 3D reconstruction}. In \bibinfo{booktitle}{\emph{Pattern Recognition and Image Analysis}}. Springer, \bibinfo{pages}{423--436}.
\newblock


\bibitem[Sadr et~al\mbox{.}(2003)]%
        {sadr2003role}
\bibfield{author}{\bibinfo{person}{Javid Sadr}, \bibinfo{person}{Izzat Jarudi}, {and} \bibinfo{person}{Pawan Sinha}.} \bibinfo{year}{2003}\natexlab{}.
\newblock \showarticletitle{The role of eyebrows in face recognition}.
\newblock \bibinfo{journal}{\emph{Perception}} \bibinfo{volume}{32}, \bibinfo{number}{3} (\bibinfo{year}{2003}), \bibinfo{pages}{285--293}.
\newblock


\bibitem[Saito et~al\mbox{.}(2018)]%
        {saito20183d}
\bibfield{author}{\bibinfo{person}{Shunsuke Saito}, \bibinfo{person}{Liwen Hu}, \bibinfo{person}{Chongyang Ma}, \bibinfo{person}{Hikaru Ibayashi}, \bibinfo{person}{Linjie Luo}, {and} \bibinfo{person}{Hao Li}.} \bibinfo{year}{2018}\natexlab{}.
\newblock \showarticletitle{3D hair synthesis using volumetric variational autoencoders}.
\newblock \bibinfo{journal}{\emph{ACM Transactions on Graphics (TOG)}} \bibinfo{volume}{37}, \bibinfo{number}{6} (\bibinfo{year}{2018}).
\newblock


\bibitem[Saito et~al\mbox{.}(2020)]%
        {saito2020pifuhd}
\bibfield{author}{\bibinfo{person}{Shunsuke Saito}, \bibinfo{person}{Tomas Simon}, \bibinfo{person}{Jason Saragih}, {and} \bibinfo{person}{Hanbyul Joo}.} \bibinfo{year}{2020}\natexlab{}.
\newblock \showarticletitle{Pifuhd: Multi-level pixel-aligned implicit function for high-resolution 3d human digitization}. In \bibinfo{booktitle}{\emph{Proceedings of the IEEE/CVF Conference on Computer Vision and Pattern Recognition}}. \bibinfo{pages}{84--93}.
\newblock


\bibitem[Schuster and Paliwal(1997)]%
        {schuster1997bidirectional}
\bibfield{author}{\bibinfo{person}{Mike Schuster} {and} \bibinfo{person}{Kuldip~K Paliwal}.} \bibinfo{year}{1997}\natexlab{}.
\newblock \showarticletitle{Bidirectional recurrent neural networks}.
\newblock \bibinfo{journal}{\emph{IEEE transactions on Signal Processing}} \bibinfo{volume}{45}, \bibinfo{number}{11} (\bibinfo{year}{1997}), \bibinfo{pages}{2673--2681}.
\newblock


\bibitem[Seitz et~al\mbox{.}(2006)]%
        {seitz2006comparison}
\bibfield{author}{\bibinfo{person}{Steven~M Seitz}, \bibinfo{person}{Brian Curless}, \bibinfo{person}{James Diebel}, \bibinfo{person}{Daniel Scharstein}, {and} \bibinfo{person}{Richard Szeliski}.} \bibinfo{year}{2006}\natexlab{}.
\newblock \showarticletitle{A comparison and evaluation of multi-view stereo reconstruction algorithms}. In \bibinfo{booktitle}{\emph{2006 IEEE computer society conference on computer vision and pattern recognition (CVPR'06)}}, Vol.~\bibinfo{volume}{1}. IEEE, \bibinfo{pages}{519--528}.
\newblock


\bibitem[Shen et~al\mbox{.}(2020)]%
        {shen2020deepsketchhair}
\bibfield{author}{\bibinfo{person}{Yuefan Shen}, \bibinfo{person}{Changgeng Zhang}, \bibinfo{person}{Hongbo Fu}, \bibinfo{person}{Kun Zhou}, {and} \bibinfo{person}{Youyi Zheng}.} \bibinfo{year}{2020}\natexlab{}.
\newblock \showarticletitle{Deepsketchhair: Deep sketch-based 3d hair modeling}.
\newblock \bibinfo{journal}{\emph{IEEE transactions on visualization and computer graphics}} \bibinfo{volume}{27}, \bibinfo{number}{7} (\bibinfo{year}{2020}), \bibinfo{pages}{3250--3263}.
\newblock


\bibitem[Sun et~al\mbox{.}(2021)]%
        {sun2021human}
\bibfield{author}{\bibinfo{person}{Tiancheng Sun}, \bibinfo{person}{Giljoo Nam}, \bibinfo{person}{Carlos Aliaga}, \bibinfo{person}{Christophe Hery}, {and} \bibinfo{person}{Ravi Ramamoorthi}.} \bibinfo{year}{2021}\natexlab{}.
\newblock \showarticletitle{Human hair inverse rendering using multi-view photometric data}.
\newblock  (\bibinfo{year}{2021}).
\newblock


\bibitem[Velinov et~al\mbox{.}(2018)]%
        {velinov2018appearance}
\bibfield{author}{\bibinfo{person}{Zdravko Velinov}, \bibinfo{person}{Marios Papas}, \bibinfo{person}{Derek Bradley}, \bibinfo{person}{Paulo Gotardo}, \bibinfo{person}{Parsa Mirdehghan}, \bibinfo{person}{Steve Marschner}, \bibinfo{person}{Jan Nov{\'a}k}, {and} \bibinfo{person}{Thabo Beeler}.} \bibinfo{year}{2018}\natexlab{}.
\newblock \showarticletitle{Appearance capture and modeling of human teeth}.
\newblock \bibinfo{journal}{\emph{ACM Transactions on Graphics (ToG)}} \bibinfo{volume}{37}, \bibinfo{number}{6} (\bibinfo{year}{2018}), \bibinfo{pages}{1--13}.
\newblock


\bibitem[Wang et~al\mbox{.}(2022)]%
        {wang2022effective}
\bibfield{author}{\bibinfo{person}{Luyuan Wang}, \bibinfo{person}{Hanyuan Zhang}, \bibinfo{person}{Qinjie Xiao}, \bibinfo{person}{Hao Xu}, \bibinfo{person}{Chunhua Shen}, {and} \bibinfo{person}{Xiaogang Jin}.} \bibinfo{year}{2022}\natexlab{}.
\newblock \showarticletitle{Effective Eyebrow Matting with Domain Adaptation}.
\newblock  (\bibinfo{year}{2022}).
\newblock


\bibitem[Wei et~al\mbox{.}(2005)]%
        {Wei2005modeling}
\bibfield{author}{\bibinfo{person}{Yichen Wei}, \bibinfo{person}{Eyal Ofek}, \bibinfo{person}{Long Quan}, {and} \bibinfo{person}{Heung-Yeung Shum}.} \bibinfo{year}{2005}\natexlab{}.
\newblock \showarticletitle{Modeling hair from multiple views}.
\newblock \bibinfo{journal}{\emph{ACM Transactions on Graphics(TOG)}} \bibinfo{volume}{24}, \bibinfo{number}{3} (\bibinfo{year}{2005}), \bibinfo{pages}{816--820}.
\newblock


\bibitem[Wen et~al\mbox{.}(2017)]%
        {wen2017real}
\bibfield{author}{\bibinfo{person}{Quan Wen}, \bibinfo{person}{Feng Xu}, \bibinfo{person}{Ming Lu}, {and} \bibinfo{person}{Jun-Hai Yong}.} \bibinfo{year}{2017}\natexlab{}.
\newblock \showarticletitle{Real-time 3D eyelids tracking from semantic edges}.
\newblock \bibinfo{journal}{\emph{ACM Transactions on Graphics (TOG)}} \bibinfo{volume}{36}, \bibinfo{number}{6} (\bibinfo{year}{2017}), \bibinfo{pages}{1--11}.
\newblock


\bibitem[Winberg et~al\mbox{.}(2022)]%
        {winberg2022facial}
\bibfield{author}{\bibinfo{person}{Sebastian Winberg}, \bibinfo{person}{Gaspard Zoss}, \bibinfo{person}{Prashanth Chandran}, \bibinfo{person}{Paulo Gotardo}, {and} \bibinfo{person}{Derek Bradley}.} \bibinfo{year}{2022}\natexlab{}.
\newblock \showarticletitle{Facial hair tracking for high fidelity performance capture}.
\newblock \bibinfo{journal}{\emph{ACM Transactions on Graphics (TOG)}} \bibinfo{volume}{41}, \bibinfo{number}{4} (\bibinfo{year}{2022}), \bibinfo{pages}{1--12}.
\newblock


\bibitem[Wu et~al\mbox{.}(2016)]%
        {wu2016model}
\bibfield{author}{\bibinfo{person}{Chenglei Wu}, \bibinfo{person}{Derek Bradley}, \bibinfo{person}{Pablo Garrido}, \bibinfo{person}{Michael Zollh{\"o}fer}, \bibinfo{person}{Christian Theobalt}, \bibinfo{person}{Markus~H Gross}, {and} \bibinfo{person}{Thabo Beeler}.} \bibinfo{year}{2016}\natexlab{}.
\newblock \showarticletitle{Model-based teeth reconstruction.}
\newblock \bibinfo{journal}{\emph{ACM Trans. Graph.}} \bibinfo{volume}{35}, \bibinfo{number}{6} (\bibinfo{year}{2016}), \bibinfo{pages}{220--1}.
\newblock


\bibitem[Wu et~al\mbox{.}(2022)]%
        {wu2022neuralhdhair}
\bibfield{author}{\bibinfo{person}{Keyu Wu}, \bibinfo{person}{Yifan Ye}, \bibinfo{person}{Lingchen Yang}, \bibinfo{person}{Hongbo Fu}, \bibinfo{person}{Kun Zhou}, {and} \bibinfo{person}{Youyi Zheng}.} \bibinfo{year}{2022}\natexlab{}.
\newblock \showarticletitle{NeuralHDHair: Automatic High-fidelity Hair Modeling from a Single Image Using Implicit Neural Representations}. In \bibinfo{booktitle}{\emph{Proceedings of the IEEE/CVF Conference on Computer Vision and Pattern Recognition}}. \bibinfo{pages}{1526--1535}.
\newblock


\bibitem[Xiao et~al\mbox{.}(2021)]%
        {xiao2021eyelashnet}
\bibfield{author}{\bibinfo{person}{Qinjie Xiao}, \bibinfo{person}{Hanyuan Zhang}, \bibinfo{person}{Zhaorui Zhang}, \bibinfo{person}{Yiqian Wu}, \bibinfo{person}{Luyuan Wang}, \bibinfo{person}{Xiaogang Jin}, \bibinfo{person}{Xinwei Jiang}, \bibinfo{person}{Yong-Liang Yang}, \bibinfo{person}{Tianjia Shao}, {and} \bibinfo{person}{Kun Zhou}.} \bibinfo{year}{2021}\natexlab{}.
\newblock \showarticletitle{Eyelashnet: A dataset and a baseline method for eyelash matting}.
\newblock \bibinfo{journal}{\emph{ACM Transactions on Graphics (TOG)}} \bibinfo{volume}{40}, \bibinfo{number}{6} (\bibinfo{year}{2021}), \bibinfo{pages}{1--17}.
\newblock


\bibitem[Xiu et~al\mbox{.}(2022)]%
        {xiu2022icon}
\bibfield{author}{\bibinfo{person}{Yuliang Xiu}, \bibinfo{person}{Jinlong Yang}, \bibinfo{person}{Dimitrios Tzionas}, {and} \bibinfo{person}{Michael~J. Black}.} \bibinfo{year}{2022}\natexlab{}.
\newblock \showarticletitle{{ICON}: {I}mplicit {C}lothed humans {O}btained from {N}ormals}. In \bibinfo{booktitle}{\emph{Proceedings of the IEEE/CVF Conference on Computer Vision and Pattern Recognition (CVPR)}}. \bibinfo{pages}{13296--13306}.
\newblock


\bibitem[Xu et~al\mbox{.}(2014)]%
        {xu2014dynamic}
\bibfield{author}{\bibinfo{person}{Zexiang Xu}, \bibinfo{person}{Hsiang-Tao Wu}, \bibinfo{person}{Lvdi Wang}, \bibinfo{person}{Changxi Zheng}, \bibinfo{person}{Xin Tong}, {and} \bibinfo{person}{Yue Qi}.} \bibinfo{year}{2014}\natexlab{}.
\newblock \showarticletitle{Dynamic hair capture using spacetime optimization}.
\newblock \bibinfo{journal}{\emph{ACM Transactions on Graphics (TOG)}}  \bibinfo{volume}{33} (\bibinfo{year}{2014}), \bibinfo{pages}{6}.
\newblock


\bibitem[Yan et~al\mbox{.}(2019)]%
        {yan2019perspective}
\bibfield{author}{\bibinfo{person}{Zhaoyi Yan}, \bibinfo{person}{Yuchen Yuan}, \bibinfo{person}{Wangmeng Zuo}, \bibinfo{person}{Xiao Tan}, \bibinfo{person}{Yezhen Wang}, \bibinfo{person}{Shilei Wen}, {and} \bibinfo{person}{Errui Ding}.} \bibinfo{year}{2019}\natexlab{}.
\newblock \showarticletitle{Perspective-guided convolution networks for crowd counting}. In \bibinfo{booktitle}{\emph{Proceedings of the IEEE/CVF international conference on computer vision}}. \bibinfo{pages}{952--961}.
\newblock


\bibitem[Yang et~al\mbox{.}(2020b)]%
        {yang2020facescape}
\bibfield{author}{\bibinfo{person}{Haotian Yang}, \bibinfo{person}{Hao Zhu}, \bibinfo{person}{Yanru Wang}, \bibinfo{person}{Mingkai Huang}, \bibinfo{person}{Qiu Shen}, \bibinfo{person}{Ruigang Yang}, {and} \bibinfo{person}{Xun Cao}.} \bibinfo{year}{2020}\natexlab{b}.
\newblock \showarticletitle{Facescape: a large-scale high quality 3d face dataset and detailed riggable 3d face prediction}. In \bibinfo{booktitle}{\emph{Proceedings of the IEEE/CVF Conference on Computer Vision and Pattern Recognition}}. \bibinfo{pages}{601--610}.
\newblock


\bibitem[Yang et~al\mbox{.}(2020a)]%
        {lingchen2020iorthopredictor}
\bibfield{author}{\bibinfo{person}{Lingchen Yang}, \bibinfo{person}{Zefeng Shi}, \bibinfo{person}{Wu Yiqian}, \bibinfo{person}{Xiang Li}, \bibinfo{person}{Kun Zhou}, \bibinfo{person}{Hongbo Fu}, {and} \bibinfo{person}{Youyi Zheng}.} \bibinfo{year}{2020}\natexlab{a}.
\newblock \showarticletitle{iOrthoPredictor: model-guided deep prediction of teeth alignment}.
\newblock \bibinfo{journal}{\emph{ACM Transactions on Graphics}} \bibinfo{volume}{39}, \bibinfo{number}{6} (\bibinfo{year}{2020}), \bibinfo{pages}{216}.
\newblock


\bibitem[Yang et~al\mbox{.}(2019)]%
        {yang2019dynamic}
\bibfield{author}{\bibinfo{person}{Lingchen Yang}, \bibinfo{person}{Zefeng Shi}, \bibinfo{person}{Youyi Zheng}, {and} \bibinfo{person}{Kun Zhou}.} \bibinfo{year}{2019}\natexlab{}.
\newblock \showarticletitle{Dynamic hair modeling from monocular videos using deep neural networks}.
\newblock \bibinfo{journal}{\emph{ACM Transactions on Graphics (TOG)}} \bibinfo{volume}{38}, \bibinfo{number}{6} (\bibinfo{year}{2019}), \bibinfo{pages}{1--12}.
\newblock


\bibitem[Zhang et~al\mbox{.}(2022)]%
        {zhang2022implicit}
\bibfield{author}{\bibinfo{person}{Congyi Zhang}, \bibinfo{person}{Mohamed Elgharib}, \bibinfo{person}{Gereon Fox}, \bibinfo{person}{Min Gu}, \bibinfo{person}{Christian Theobalt}, {and} \bibinfo{person}{Wenping Wang}.} \bibinfo{year}{2022}\natexlab{}.
\newblock \showarticletitle{An Implicit Parametric Morphable Dental Model}.
\newblock \bibinfo{journal}{\emph{ACM Transactions on Graphics (TOG)}} \bibinfo{volume}{41}, \bibinfo{number}{6} (\bibinfo{year}{2022}), \bibinfo{pages}{1--13}.
\newblock


\bibitem[Zhang et~al\mbox{.}(2017)]%
        {zhang2017data}
\bibfield{author}{\bibinfo{person}{Meng Zhang}, \bibinfo{person}{Menglei Chai}, \bibinfo{person}{Hongzhi Wu}, \bibinfo{person}{Hao Yang}, {and} \bibinfo{person}{Kun Zhou}.} \bibinfo{year}{2017}\natexlab{}.
\newblock \showarticletitle{A data-driven approach to four-view image-based hair modeling.}
\newblock \bibinfo{journal}{\emph{ACM Trans. Graph.}} \bibinfo{volume}{36}, \bibinfo{number}{4} (\bibinfo{year}{2017}), \bibinfo{pages}{156--1}.
\newblock


\bibitem[Zhang et~al\mbox{.}(2018)]%
        {zhang2018modeling}
\bibfield{author}{\bibinfo{person}{Meng Zhang}, \bibinfo{person}{Pan Wu}, \bibinfo{person}{Hongzhi Wu}, \bibinfo{person}{Yanlin Weng}, \bibinfo{person}{Youyi Zheng}, {and} \bibinfo{person}{Kun Zhou}.} \bibinfo{year}{2018}\natexlab{}.
\newblock \showarticletitle{Modeling hair from an rgb-d camera}.
\newblock \bibinfo{journal}{\emph{ACM Transactions on Graphics (TOG)}} \bibinfo{volume}{37}, \bibinfo{number}{6} (\bibinfo{year}{2018}).
\newblock


\bibitem[Zhang and Zheng(2019)]%
        {zhang2019hair}
\bibfield{author}{\bibinfo{person}{Meng Zhang} {and} \bibinfo{person}{Youyi Zheng}.} \bibinfo{year}{2019}\natexlab{}.
\newblock \showarticletitle{Hair-GAN: Recovering 3D hair structure from a single image using generative adversarial networks}.
\newblock \bibinfo{journal}{\emph{Visual Informatics}} \bibinfo{volume}{3}, \bibinfo{number}{2} (\bibinfo{year}{2019}), \bibinfo{pages}{102--112}.
\newblock


\bibitem[Zheng et~al\mbox{.}(2023)]%
        {zheng2023hairstep}
\bibfield{author}{\bibinfo{person}{Yujian Zheng}, \bibinfo{person}{Zirong Jin}, \bibinfo{person}{Moran Li}, \bibinfo{person}{Haibin Huang}, \bibinfo{person}{Chongyang Ma}, \bibinfo{person}{Shuguang Cui}, {and} \bibinfo{person}{Xiaoguang Han}.} \bibinfo{year}{2023}\natexlab{}.
\newblock \showarticletitle{Hairstep: Transfer synthetic to real using strand and depth maps for single-view 3d hair modeling}. In \bibinfo{booktitle}{\emph{Proceedings of the IEEE/CVF Conference on Computer Vision and Pattern Recognition}}. \bibinfo{pages}{12726--12735}.
\newblock


\bibitem[Zheng et~al\mbox{.}(2021)]%
        {zheng2021pamir}
\bibfield{author}{\bibinfo{person}{Zerong Zheng}, \bibinfo{person}{Tao Yu}, \bibinfo{person}{Yebin Liu}, {and} \bibinfo{person}{Qionghai Dai}.} \bibinfo{year}{2021}\natexlab{}.
\newblock \showarticletitle{Pamir: Parametric model-conditioned implicit representation for image-based human reconstruction}.
\newblock \bibinfo{journal}{\emph{IEEE transactions on pattern analysis and machine intelligence}} \bibinfo{volume}{44}, \bibinfo{number}{6} (\bibinfo{year}{2021}), \bibinfo{pages}{3170--3184}.
\newblock


\bibitem[Zhou et~al\mbox{.}(2018)]%
        {zhou2018hairnet}
\bibfield{author}{\bibinfo{person}{Yi Zhou}, \bibinfo{person}{Liwen Hu}, \bibinfo{person}{Jun Xing}, \bibinfo{person}{Weikai Chen}, \bibinfo{person}{Han-Wei Kung}, \bibinfo{person}{Xin Tong}, {and} \bibinfo{person}{Hao Li}.} \bibinfo{year}{2018}\natexlab{}.
\newblock \showarticletitle{Hairnet: Single-view hair reconstruction using convolutional neural networks}. In \bibinfo{booktitle}{\emph{Proceedings of the European Conference on Computer Vision (ECCV)}}. \bibinfo{pages}{235--251}.
\newblock


\bibitem[Zhu and Bridson(2005)]%
        {zhu2005animating}
\bibfield{author}{\bibinfo{person}{Yongning Zhu} {and} \bibinfo{person}{Robert Bridson}.} \bibinfo{year}{2005}\natexlab{}.
\newblock \showarticletitle{Animating sand as a fluid}.
\newblock \bibinfo{journal}{\emph{ACM Transactions on Graphics (TOG)}} \bibinfo{volume}{24}, \bibinfo{number}{3} (\bibinfo{year}{2005}), \bibinfo{pages}{965--972}.
\newblock


\bibitem[Zielonka et~al\mbox{.}(2022)]%
        {zielonka2022towards}
\bibfield{author}{\bibinfo{person}{Wojciech Zielonka}, \bibinfo{person}{Timo Bolkart}, {and} \bibinfo{person}{Justus Thies}.} \bibinfo{year}{2022}\natexlab{}.
\newblock \showarticletitle{Towards metrical reconstruction of human faces}. In \bibinfo{booktitle}{\emph{Computer Vision--ECCV 2022: 17th European Conference, Tel Aviv, Israel, October 23--27, 2022, Proceedings, Part XIII}}. Springer, \bibinfo{pages}{250--269}.
\newblock


\bibitem[Zoss et~al\mbox{.}(2019)]%
        {zoss2019accurate}
\bibfield{author}{\bibinfo{person}{Gaspard Zoss}, \bibinfo{person}{Thabo Beeler}, \bibinfo{person}{Markus Gross}, {and} \bibinfo{person}{Derek Bradley}.} \bibinfo{year}{2019}\natexlab{}.
\newblock \showarticletitle{Accurate markerless jaw tracking for facial performance capture}.
\newblock \bibinfo{journal}{\emph{ACM Transactions on Graphics (TOG)}} \bibinfo{volume}{38}, \bibinfo{number}{4} (\bibinfo{year}{2019}), \bibinfo{pages}{1--8}.
\newblock


\bibitem[Zoss et~al\mbox{.}(2018)]%
        {zoss2018empirical}
\bibfield{author}{\bibinfo{person}{Gaspard Zoss}, \bibinfo{person}{Derek Bradley}, \bibinfo{person}{Pascal B{\'e}rard}, {and} \bibinfo{person}{Thabo Beeler}.} \bibinfo{year}{2018}\natexlab{}.
\newblock \showarticletitle{An empirical rig for jaw animation}.
\newblock \bibinfo{journal}{\emph{ACM Transactions on Graphics (TOG)}} \bibinfo{volume}{37}, \bibinfo{number}{4} (\bibinfo{year}{2018}), \bibinfo{pages}{1--12}.
\newblock


\end{thebibliography}
